\documentclass[letterpaper]{article} 
\usepackage{aaai23}  
\usepackage{times}  
\usepackage{helvet}  
\usepackage{courier}  
\usepackage[hyphens]{url}  
\usepackage{graphicx} 
\usepackage{natbib}  
\usepackage{caption} 
\usepackage{algorithm}
\usepackage{algorithmic}

\usepackage{newfloat}
\usepackage{listings}
\DeclareCaptionStyle{ruled}{labelfont=normalfont,labelsep=colon,strut=off} 
\floatstyle{ruled}
\newfloat{listing}{tb}{lst}{}
\floatname{listing}{Listing}
\usepackage{times}
\usepackage{epsfig}
\usepackage{graphicx}
\usepackage{amsmath}
\usepackage{amssymb}
\usepackage{booktabs}
\usepackage{paralist}
\usepackage{multirow}
\usepackage{tabularx}
\usepackage{bbm}

\DeclareMathOperator*{\argmax}{argmax}
\DeclareMathOperator*{\argmin}{argmin}

\title{Inference Time Evidences of Adversarial Attacks for Forensic on Transformers}
\author{
    Hugo Lemarchant,\textsuperscript{\rm 1}
    Liangzi Li,\textsuperscript{\rm 1}
    Yiming Qian,\textsuperscript{\rm 1}
    Yuta Nakashima,\textsuperscript{\rm 1}
    Hajime Nagahara
}
\affiliations{
    \textsuperscript{\rm 1}Osaka University\\
    hugo@is.ids.osaka-u.ac.jp, li@ids.osaka-u.ac.jp, yimingqian@ids.osaka-u.ac.jp, n-yuta@ids.osaka-u.ac.jp, nagahara@ids.osaka-u.ac.jp
}

\usepackage{bibentry}

\begin{document}

\maketitle

\begin{abstract}
Vision Transformers (ViTs) are becoming a very popular paradigm for vision tasks as they achieve state-of-the-art performance on image classification.
However, although early works implied that this network structure had increased robustness against adversarial attacks, some works argue ViTs are still vulnerable.
This paper presents our first attempt toward detecting adversarial attacks during inference time using the network's input and outputs as well as latent features.
We design four quantifications (or derivatives) of input, output, and latent vectors of ViT-based models that provide a signature of the inference, which could be beneficial for the attack detection, and empirically study their behavior over clean samples and adversarial samples. 
The results demonstrate that the quantifications from input (images) and output (posterior probabilities) are promising for distinguishing clean and adversarial samples, while latent vectors offer less discriminative power, though they give some insights on how adversarial perturbations work. 
\end{abstract}

\section{Introduction}
All deep learning models used nowadays are vulnerable to adversarial attacks \cite{goodfellow2013intriguing, NaseerIntriguingViT2021}, especially in the domain of image classification.
This is inherent to the fact that gradient computation is required for training \cite{SGDKeifer1952}, which can also be used to optimize adversarial noise instead of model weights.
Current research efforts around adversarial attacks are often directed toward exploring why such attacks exist, where the vulnerability comes from \cite{goodfellow2013intriguing}, and even how to make the attack more efficient \cite{Biggio_2018}.

The situation is the same for recently-emerged Vision Transformers (ViTs) \cite{dosovitskiy2021image}. By incorporating the self-attention mechanism, some portions of the network became more understandable, but the vulnerability inherent to a stack of a simple but large number of arithmetic computations can still be exploited for adversarial attacks.
Many papers ask ``Could ViTs be more robust than CNN?'' or ``To which attack are they more robust?'' \cite{benz2021robustness,Bai2021Are,raghu2021vision,shao2021adversarial,Paul2021vision,MahmoodOnTheRobust2021}.
These studies gave us some insights into how ViTs perform against attacks, sometimes compared to other families of networks. Yet, whether or not ViTs can offer fundamental ways to avoid vulnerability is still an open question. For now, we may need to acknowledge that getting rid of this vulnerability is a hard problem.

Adversarial attacks may have been an armchair problem, but the circumstance may be changing rapidly. Deepfakes, for example, have become an actual threat by counterfeiting authorities' statements \cite{suwajanakorn2017synthesizing}, whereas researchers are working toward detecting such deepfakes using neural networks \cite{albahar2019deepfakes}, playing a cat-and-mouse game. Adversarial attacks may have the potential to become a game-changer by facilitating the deepfake camp to fool deepfake detectors.

A potential alternative to inventing a mechanism to robustify ViTs is providing some ways to detect adversarial attacks.
Closely related studies \cite{raghu2021vision, CintasAttactDetect2020} presented for the first time feature representation \cite{kornblith2019similarity} and attention profile analysis \cite{raghu2021vision} on ViTs, applied for comparisons against CNNs and between pre-trained and trained-from-scratch ViTs.

We propose to extend these studies exploring \emph{inference time signatures of adversarial attacks} that potentially imply their presence in the input, to refine our understanding of the reactive part of a network.
This work is a preliminary to a forensic exercise, where we wish to collect evidence of an attack given a suspicious input.

More specifically, for vanilla ViTs and tokens-to-token ViTs (T2T-ViTs) \cite{dosovitskiy2021image,yuan2021tokenstotoken} under various attacks, we show whether the input and the latent vectors (including self-attentions) have any signatures of the attack at multiple levels, including (i) the input itself and its frequency decomposition, (ii) output vector, (iii) attention profile \cite{raghu2021vision}, (iv) the feature similarity given by centered kernel alignment (CKA) \cite{raghu2021vision, kornblith2019similarity}.

Our findings are summarized as follows:
\begin{itemize}
    \item \textbf{Input and output of the model tell a lot.} Low-frequency components in adversarialy perturbed images carry more energy. Also, the entropy of posterior probability over all classes (for classification tasks) increases. These results support the results in \cite{UelwerLearning2021,GuptaAdv2020} and show that adversarial attacks are relatively visible outside the model, which may make designing a detector feasible.
    \item \textbf{Latent vectors are not discriminative but explanatory.} Some of our quantifications are based on the model's latent vectors, and they have lower discrimination power to detect attacks. However, they provide more complete information about which features are more sensitive and how they react to attacks. This explainability makes these quantifications valuable.
    \item \textbf{Perturbations may spot the acquired structure of the model.} It is known that ViTs have some inductive bias, which renders block-like structure in the CKA similarity matrix. We empirically show that the latent vectors at the ``boundaries'' of this structure react to adversarial perturbations. This fact can facilitate an attack detector tailored for a certain model.
\end{itemize}

This makes our contribution two-fold: First, we present and evaluate the first draft of quantifications that potentially show signatures of adversarial attacks on ViTs. Second, we present intriguing internal behavior that may realize a more robust network.

\section{Related Work}
Many recent works demonstrate the robustness of the ViTs and their derivatives, such as \cite{benz2021robustness,Bai2021Are,shao2021adversarial,Paul2021vision,towardsRobustMao2021}.
However, these works focus only on evaluating how much robustness gain can be obtained from ViTs compared to CNNs \cite{MahmoodOnTheRobust2021}, or how to robustify the architecture \cite{towardsRobustMao2021}.
We will present in the following sections the literature review on the robustness of ViTs in image classification tasks and on the feature representation within the Transformers to rationalize our choice of working on it to find signatures of attacks.

\subsection{Adversarial attacks}

The idea of adversarial attacks emerged \cite{goodfellow2013intriguing} soon after the emergence of deep neural networks and has now become a well-known research topic with many variations to fool different architectures \cite{fu2022patch} and knowledge levels, such as white-box and black-box \cite{Biggio_2018} attacks, warning the potential vulnerability of the current machine learning trend.

On the defense side, most of the efforts towards better robustness improvement are made for architectural modifications \cite{towardsRobustMao2021} and input scrambling \cite{zantedeschi2017efficient}.
Evaluating the robustness of a specific model is often limited to input analysis \cite{carlini2017adversarial, hendrycks2016early}, with limited attention given to the network behavior during inference. Other techniques are using pre-processing to reduce the impact of the following network basing their strategy on transformations \cite{li2021learning}, bringing back any perturbation into the training distribution, or blending/blurring the local information to reduce the chance of successful attacks. While those techniques have proven to be successful, their application is often limited to several architecture families or attack families, and  they may impair the accuracy of the models.

Not knowing the internal representations made during inference for sure hurts the possibilities to detect any attack.
This idea motivates us to explore the internal structure of models in the inference time.

\subsection{Robustness of ViTs}

ViTs introduced in \cite{dosovitskiy2021image} use the original design of the Transformer \cite{vaswani2017attention,devlin2019bert} but decompose the picture into patches as a sentence is decomposed into words.
One drawback of ViTs is the computationally demanding pre-training stage required to maximize the generalization of the network, but it is addressed with various architectural tricks such as T2T \cite{yuan2021tokenstotoken} with no significant drawback in clean accuracy \cite{shao2021adversarial}, but in their experiment they show up to 40\% of adversarial accuracy loss.
This architecture scored state of the art on image classification benchmark and in real-world applications \cite{han2022survey}.
The self attention mechanism at heart of the ViT family seems to helps in the robustness to out-of-distribution samples \cite{wang2022can,shao2021adversarial}.
However those papers and attack specific ones such as \cite{fu2022patch} show that the ViTs actually trade comparable results depending on the task at hand.
We are currently heavily evaluating the adversarial robustness on standard attack sets, comprised of traditional attacks mainly developed on Fully Connected NNs and CNNs.
As shown in \cite{fu2022patch}, it is possible to specifically attack self attention basic operation with success.
In sight of that, we think that acknowledging the robustness of ViTs in all cases is dangerously optimistic, and studying their adversarial vulnerability and behavior should be also seriously considered.

\subsection{Feature Representations in ViTs}

The authors of \cite{raghu2021vision} discovered two characteristics in the features of the ViTs that differs from the CNN.
They firstly present an inductive bias learned by the networks undergoing long pretraining and they show that those networks learn how to attend to global and local features, while those simply trained from scratch are limited to global features.
The ViTs "learn" how to attend to local features, when the CNNs have this implemented in the architecture.
Then they proceed in showing the appearance more monolithic of the CKA matrix computed on ViTs compared to the one of the CNNs.
Indeed the later one did show two blocks, while the ViT had only one, meaning the features extracted from the beginning were forwarded to the next block in a very consistent way.
The stability of the feature representation throughout the network is promising for attack detection. 
Additionally, self-supervised applications of the ViTs such as \cite{caron2021emerging} show that certain condition of training favor the emergence of semantically meaningful features right out of the network.
This echoes to the final observation of \cite{raghu2021vision} where they expose the self-location properties of the ViTs heads.
While we do not study self-supervised trained models, one can ask if the attacks affect the capability of a Self-Attention head to focus on locally relevant information.

\section{Inference Time Signatures of Attacks}
\label{sec:sig}

Due to the complexity of deep neural networks, identifying signatures of adversarial attacks is rather exploratory. 
We introduce here five derivatives of an input, a batch of inputs, or their latent vectors that potentially exhibit reactions of a Transformer network to adversarial attacks.
These derivatives are not our new proposal, but we re-purpose the original intent of the derivatives for single-sample adversarial attack detection.

\subsection{Signatures in the frequency domain}

Previous studies have paid attention to the energy spectrum of the adversarial noises demonstrating that ViTs require a wider-spread spectrum for successful attacks \cite{shao2021adversarial} while CNNs are more vulnerable to high-frequency although they are still crucial for the attack on ViT \cite{Paul2021vision}. These results suggest that adversarial attacks tailored for ViTs affect a wider portion of the input in the frequency domain. Although the original image and adversarial noise are inseparable in our inference time detection scenario, the energy spectrum may provide some clues on the pattern of attacks. 

We thus compute the discrete cosine transform (DCT) of the input.
According to the previous results, we will be particularly attentive to the frequency variations in input sample $x$, either a clean image $x = o$ or attacked image $x = o + \eta$, where $o$ is the original image and $\eta$ is the adversarial noise, that is, attacked images may have more lower-frequency components carrying more energy compared to the original images. 
For quantification, we propose the \emph{frequency ratio}, which gives the ratio of high-frequency components over the low-frequency components of the energy spectrum. Let $\text{DCT}_{ij}(x)$ denote the $(i,j)$-th energy component of $x$'s DCT. Frequency ratio $\text{FR}$ is defined as:
\begin{equation}
    \text{FR}(x) = \frac {\sum_{(i,j) \in \text{HF}} \text{DCT}_{ij}(x)}{\sum_{(i,j) \in \text{LF}} \text{DCT}_{ij}(x)},
\end{equation}
where $\text{HF}$ and $\text{LF}$ are the sets of indexes of high- and low-frequency components, given using threshold $\phi$ by $\text{HF} = \{(i,j)|i + j \geq \phi\}$ and $\text{LF} = \{(i,j)|i + j < \phi\}$.

\subsection{Signatures in posterior probabilities}

The output $f(x)$ (after softmax) of any classifier is usually interpreted as the posterior probabilities for respective classes of the task. 
Adversarial attacks, in turn, try to reduce the probability of the ground-truth class (or increase the probabilities of the other classes). 
A naive deduction from this argument is that the \emph{posterior entropy} $\text{PH}$ can be a basic signature of adversarial attacks. 
\begin{equation}
\text{PH}(x) = -\sum_{k=1}^K f_k(x) \log f_k(x),
\label{Entropy}
\end{equation}
where $f_k(x)$ is the $k$-th element of $f(x)$ for $k = 1,\dots,K$.

The idea behind this quantification is that adversarial noises well-tuned for the model and the visual quality may hit the almost same probability masses for the ground-truth class and the second-top class, leading to higher $\text{PH}$.

\subsection{Signatures in attention distances}
\label{sec:attdist}

CNNs have embedded inductive bias in their structure that constrains possible dependency among pixels only to their neighbors. 
Literature \cite{raghu2021vision} has shown that vanilla ViTs learn to attend neighboring patches when extensively pre-trained with a large dataset, such as JFT-300M \cite{sun2017revisiting}, but still ViTs have an ability to attend wherever necessary for the classification task. 
Meanwhile, ViTs are typically more robust to adversarial attacks compared to CNNs. 
This contrast can imply that texture features (or local dependency) is prone to successful attacks while shape features (or global dependency) give better robustness. We argue that the locality of dependency can be a key factor for robustness and are curious about the validity of this statement under adversarial attacks. 

For ViTs, the locality of dependency (beyond patches) is encoded in the self-attention mechanism; that is, attention from a certain patch to only its neighbors suggests local dependency, while the attention spanning all over the input image suggests global dependency. We thus define the \emph{attention distance} as the average of the distances in pixel between a certain patch and any other patch weighted by the attention value associated to this pair of patches.

\begin{figure}[t]
\begin{center}
\includegraphics[width=\linewidth]{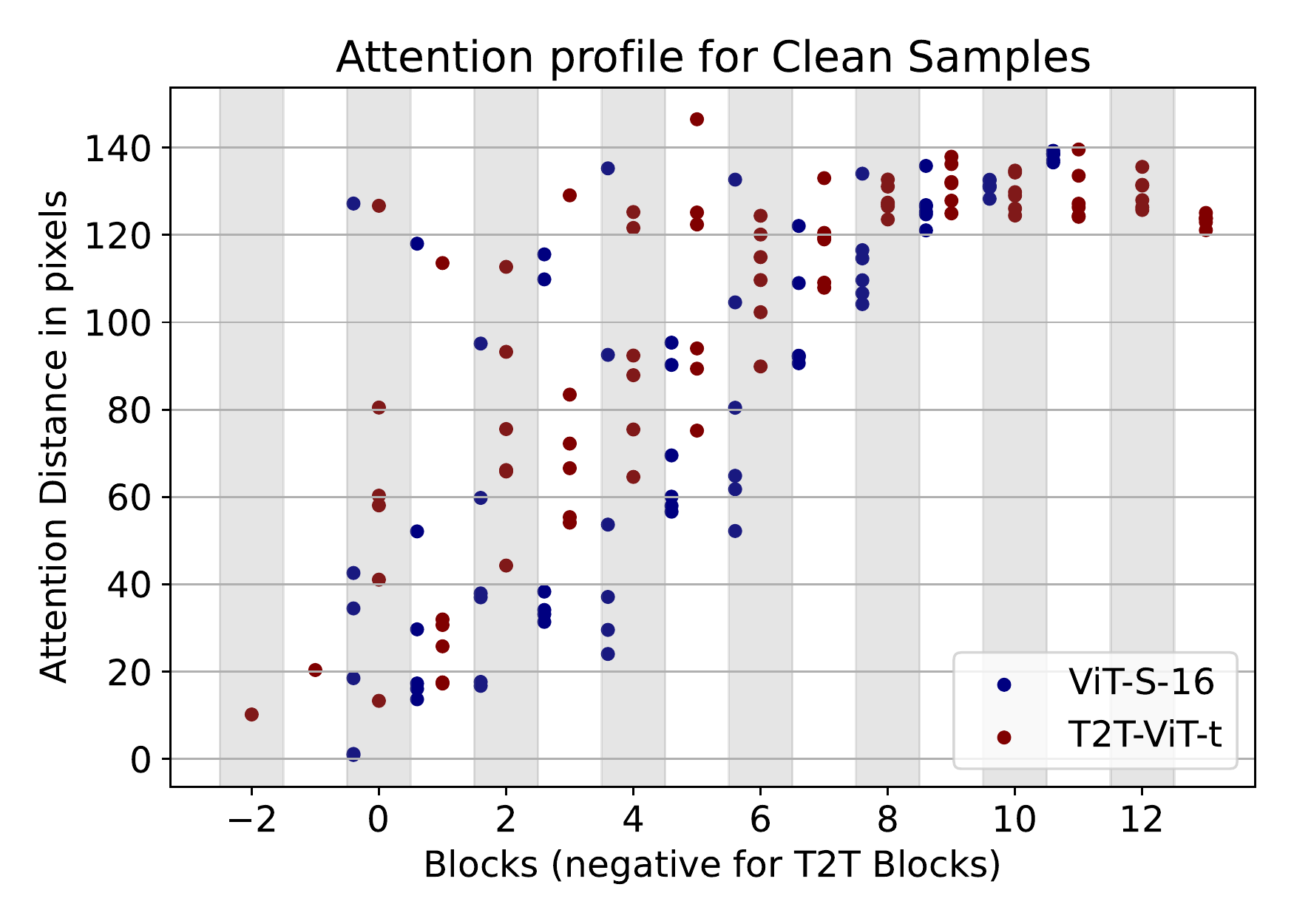}
\end{center}
\vspace{-0.5cm}
\caption{Example attention profiles of two models ViT-S-16 and T2T-ViT-t evaluated over the ImageNet test dataset. Each block has multiple dots, which are $\text{AD}(x)$ for respective self-attention heads, averaged over all $x$'s in the dataset.}
\label{fig:Att_profile_ViT_T2T}
\end{figure}

Specifically, let $A_{ij}(x)$ be the attention between patches $i$ and $j$ of a certain Transformer block and a certain self-attention head for input $x$, and $d_{ij}$ the distance between the centers of patches $i$ and $j$. 
Attention distance $\text{AD}$ for this self-attention head is given by
\begin{equation}
    \text{AD}(x) = \frac{\sum_{ij} A_{ij}(x) \cdot d_{ij}} {\sum_{ij}A_{ij}(x)},
\label{HeadDist}
\end{equation}
where the summations are computed for all patch pairs. 

We collectively refer to $\text{AD}(x)$'s for all Transformer blocks and self-attention heads as the \emph{attention profile} (AP). Figure \ref{fig:Att_profile_ViT_T2T} shows example APs for ViT and T2T-ViT models, evaluated over clean input samples (i.e., $x = o$).
For readability, we arrange them in a block-wise manner, where the negative block indexes are for the T2T blocks.
APs can be computed for a set of samples by averaging $\text{AD}(x)$ over $x$ in the set. 

An AP consists of multiple ADs, but it may be convenient to show a single-value summary of AP for a consistent display with other quantifications. Let $\bar{\text{AD}}^\text{ref}_{bh}$ and $\text{AD}_{bh}(x)$ be the mean AD computed for the set of clean samples and AD computed for input $x$. We define summary $S_\text{AP}(x)$ as the sum of their absolute differences, i.e.,
\begin{equation}
    S_\text{AP}(x) = \sum_{b,h} |\bar{\text{AD}}^\text{ref}_{bh} - \text{AD}_{bh}(x)|,
\end{equation}
where the summation is computed over all blocks and heads of the model. 

We can generate the attention profile for a trained model over a set of clean input samples, which gives reference distributions of attention distances. Comparison between these reference distributions and a (set of) suspicious input may provide some ideas on the presence of adversarial attacks.

\subsection{Structures in latent representations}

CKA \cite{raghu2021vision} is used for studying the representation structure in a model, by giving layer-to-layer similarities in that model. Analyzing latent representations within a model allows for identifying different general functions throughout its architecture.
Adversarial attacks may break such functions as they exploit some patterns in the learned parameters, which results in unusual responses. It should be noted that, although we aim to discover inference time signatures of adversarial attacks, CKA is a statistic and is computed only for a set of samples but not for a single sample; therefore, this signature is not always usable.

CKA takes as input $H_i \in \mathbb{R}^{m \times c}$ and $H_j \in \mathbb{R}^{m \times c'}$, which are latent vectors (or activation) at the $i$-th and $j$-th layers ($i, j = 1,\dots,l$) with the dimensionalities of $c_i$ and $c_j$, respectively, obtained for the same $m$ input samples. Letting $K_i = H_i H_i^\top$ and $K_j = H_j H_j^\top$ denote the Gram matrices, CKA is given by:
\begin{equation}
    \text{CKA}_{ij} = \frac{\text{HSIC}(K_i, K_j)}{\sqrt{\text{HSIC}(K_i,K_i) \; \text{HSIC}(K_j,K_j)}}
\end{equation}
where $\text{HSIC}$ is the Hilbert-Schmidt independence criterion. 
Given the centering matrix\footnote{Here we omit the indices $i$ and $j$ for notation simplicity.} $H_c = I_c-\frac{1}{c} \mathbbm{1}_c$ with $I_c$ being the $c \times c$ diagonal matrix and $\mathbbm{1}_c$ being the $c \times c$ matrix whose elements are all 1, the centered Gram matrix for $K$ is computed by $K' = H_c K H_c$. This operation is applied to both $K_i$ and $K_j$ to obtain $K_i'$ and $K_j'$.  $\text{HSIC}$ is given as the similarity between these centered Gram matrices, computed by 
\begin{equation}
\text{HSIC}(K_i, K_j) = \text{vec}(K_i')^\top \text{vec}(K_j')/(m - 1)^2,
\end{equation}
where $\text{vec}(\cdot)$ is the vectorization of a matrix.
CKA is usually computed over the entire test set because of the statistical nature of the metric.
Hence, we compute CKA for the minimum batch size (i.e. $m = 4$).

CKA is a similarity between an arbitrary pair of layers, resulting in a similarity matrix. But this is not always handy for our purpose of finding signatures of adversarial attacks. We reduce the similarity matrix down to a single metric for compacting the observations. Also to highlight the reactions to adversarial attacks, we compute the difference between CKAs computed for clean and attacked samples. 

Specifically, let $M^\text{ref} \in \mathbb{R}^{l\times l}$ denote the reference CKA similarity matrix computed for clean samples in the entire test set (i.e., the $(i,j)$-th element of $M^\text{ref}$ is $\text{CKA}_{ij}$ computed for clean samples). Also, let $\bar{M}$ denote the mean of CKA similarity matrices of the same model computed for random mini-batches with the minimum batch size and of samples under the attack of interest (or clean samples), i.e.,
\begin{equation}
    \bar{M} = \frac{1}{|\mathcal{B}|} \sum_{B \in \mathcal{B}} M(B),
\end{equation}
where $\mathcal{B}$ is the set of aforementioned mini-batches and $M(B)$ is the CKA similarity matrix computed for $B \in \mathcal{B}$.
To highlight the changes, we compute the absolute difference $D$ between $M^\text{ref}$ and $\bar{M}$ obtained with samples under the attack of interest, where its $(i,j)$-th element $D_{ij}$ is given by 
\begin{equation}
    D_{ij} = |M^\text{ref}_{ij} - \bar{M}_{ij}|.
\end{equation}
The summary $S_\text{CKA}$ is defined as the sum of these differences, i.e.,
\begin{equation}
    S_\text{CKA} = \sum_{ij} D_{ij}.
\end{equation}

\section{Experimental Setup}

\subsection{Models}
We choose two major architectural families that cover various paradigms of ViTs.
The first family is the vanilla ViT \cite{dosovitskiy2021image}, which comes with 3 variants: ViT/T-16, ViT/B-32, and ViT/B-16.
This allows us to draw more general conclusions as they consist only of Transformer blocks, which are the main building block of the majority of the entire ViT family.
We also investigate T2T-ViTs \cite{yuan2021tokenstotoken} with two variants: T2T-ViT performer \cite{ChenViT-P2022, ChororethinhkPerf2020} and the original T2T-ViT transformer.
The T2T-ViT family addresses the need for extensive pre-training on large datasets, which is inevitable for the vanilla ViTs. By introducing a CNN-like structure, this family also happens addressing the inductive bias \cite{raghu2021vision}, bringing a T2T-ViT model trained from scratch over ImageNet with an attention profile comparable to fully pre-trained ViTs (Figure \ref{fig:Att_profile_ViT_T2T}).
Evaluation on this family provides some insights about adversarial attacks over architectural variations like hierarchical designs in ViT \cite{ze2021SWIN} and  variations without pre-training \cite{Touvron2020DEIT}.

\begin{table}
  \begin{center}
    {\small{
\begin{tabular}{lrrrr}
\toprule
 Model & Encoding & Depth & \# heads & \# params.\\
\midrule
 ViT-T/16 & Conv. & 12 & 3 & 5.7M\\
 ViT-S/32 & Conv. & 12 & 6 & 22M\\
 ViT-S/16 & Conv. & 12 & 6 & 22.6M\\
 \hline
 T2T-ViT-p & Performer & 14 & 6 & 21.5M\\
 T2T-ViT-t & Transformer & 14 & 6 & 21.5M\\
\bottomrule
\end{tabular}
}}
\end{center}
\caption{Summary of ViT and T2T-ViT models used in our experiments.}
\label{tab:model-tab}
\end{table}

Table \ref{tab:model-tab} shows a summary of models we used. For vanilla ViT variants, we used models pre-trained on JFT-300M.\footnote{{\scriptsize\url{https://github.com/rwightman/pytorch-image-models}} and {\scriptsize\url{https://github.com/yitu-opensource/T2T-ViT}}} All models are trained (or fine-tuned) on the ImageNet training set \cite{SteinerHowToTrain2021}. 
For legibility, we present the results for ViT-S/16 and T2T-ViT-t; the other models' results are presented in the supplementary material.

\subsection{Dataset}
Our study is conducted over the ImageNet dataset, which is large enough to train the aforementioned models. The size of input images is $224\times 224$, which allows us to optimize adversarial perturbations in a reasonable time.
To perform our attacks, we will be using the ImageNet-1K test set of 50k samples.
To unload the computational burden, we sampled one out of five samples from the set, giving us 10k samples, which is still representative of the dataset.

\subsection{Adversarial Attacks}
\label{sec:adv_attack}

We evaluated the robustness under popular adversarial attacks, i.e., the fast gradient sign method (FSGM) \cite{goodfellow2014explaining}, projected gradient descent (PGD) \cite{TowardsMadry2017}, and Carlini and Wagner (C\&W) \cite{CWTowardsEval2016}. 

FSGM and PGD are $l_\infty$ attacks, where attacked image $x$ for original image $o$ is given by 
\begin{equation}
    x = \argmax_{x'}\mathcal{L}(x', y)\;\; \mathrm{s.t.}\;\; \|x' - o\|_{\infty} \leq \epsilon,
\end{equation}
where $\mathcal{L}$ is the loss function for the task (typically, softmax cross-entropy) and $y$ is the ground-truth label.
This is an untargeted attack, in which the attack is successful when the perturbation changes the model's prediction.
The above optimization problem is intractable because of the $l_\infty$ constraint. 
FGSM approximates the optimization problem by
\begin{equation}
    x = o + \epsilon\; \text{sign}(\nabla_o \mathcal{L} (o, y)),
\end{equation}
where $\nabla_o \mathcal{L}$ is the gradient of $\mathcal{L}$ at $o$.
This is a quick but rough approximation. 

PGD is one of FGSM's extensions, which iteratively performs FGSM. Letting $x_t$ denote the attacked image after $t$ iterations, PGD computes $x_{t+1}$ by
\begin{equation}
    x_{t+1} = \text{clip}_{[o - \epsilon, o + \epsilon]}(x_t + \alpha \; \text{sign}(\nabla_x\mathcal{L}(x_{t}, y))),
\end{equation}
where $\text{clip}_{[o - \epsilon, o + \epsilon]}(\cdot)$ is to clip the value within $[o - \epsilon, o + \epsilon]$, $\alpha$ is a predefined small value, and $x_0 = o$.
This iterative method takes advantage of multiple small steps, controlled by $\alpha$, to refine the attack. Iteration enables the attack to be more efficient than FGSM, so that a much smaller $\epsilon$ radius enables the same level of perturbation as FGSM.
However, PDG is more computationally demanding; the computation cost is linear with the number of iterations.
We will use PGD for 40 iterations with $\alpha = 0.025$.

In our experiments, we use $\epsilon \in \{0.031, 0.062\}$ for FGSM and $\epsilon \in \{0.001, 0.003, 0.005, 0.01\}$ for PGD.

C\&W jointly optimizes the distortion introduced by the perturbation in the $l_2$ distance and the performance of the attack.
The method quantifies the performance of the attack by
\begin{equation}
    Q(x)= \mathrm{max}\left[f_y(x) - \max_{y' \neq y} f_{y'}(x), -\kappa\right],
\end{equation}
where we use the implementation\footnote{\scriptsize\url{https://github.com/Harry24k/adversarial-attacks-pytorch}} of the untargeted version; $f_y$ and $f_{y'}$ denote the prediction confidences for ground-truth label $y$ and another label $y'$. A negative value of $Q(x)$ means that a certain class other than the ground truth gets the highest confidence, though the confidence does not go beyond the margin $\kappa$.  

C\&W parameterizes attacked image $x$ by $w$, i.e.,
\begin{equation}
    x(w) = \frac{\tanh(w) + 1}{2},
\end{equation}
which guarantees $x$ in the range of $[0, 1]$ during the optimization.\footnote{We normalize original image $o$ to be in $[0, 1]$ as well}. The attacked image $x$ of C\&W is given by $x(w^*)$, where
\begin{equation}
    w^* = \argmin_w \| x(w) - o\|^2 + c \; Q(x(w))
\end{equation}
with $c = 1\times 10^{-4}$.

\begin{table}[t]
\caption{Accuracy of ViT and T2T models under various attacks.}
\small
\begin{tabular}{lrrrr}
\toprule
& \multicolumn{2}{c}{Top-1} & \multicolumn{2}{c}{Top-5} \\
\cmidrule(lr){2-3} \cmidrule(lr){4-5}
Attack          & ViT-S/16   & T2T-ViT-t   & ViT-S/16   & T2T-ViT-t    \\
\midrule
Clean           & 81.38      & 81.40      & 96.13      & 95.72      \\
C\&W {\scriptsize $c\! =\! 1\!\!\times\!\! 10^{-4}$}\hspace*{-3mm}    & 24.86      & 26.21      & 94.68      & 92.99      \\
PGD {\scriptsize $\epsilon\! =\! 1\!\!\times\!\! 10^{-3}$}\hspace*{-4mm}   & 51.36      & 44.37      & 86.99      & 80.79      \\
PGD {\scriptsize $\epsilon\! =\! 3\!\!\times\!\! 10^{-3}$}\hspace*{-4mm}  & 20.47      & 22.98       & 71.65      & 60.40        \\
PGD {\scriptsize $\epsilon\! =\! 5\!\!\times\!\! 10^{-3}$}\hspace*{-4mm}    & 11.36      & 17.30      & 60.75      & 49.68      \\
PGD {\scriptsize $\epsilon\! =\! 1\!\!\times\!\! 10^{-2}$}\hspace*{-4mm}   & 6.50       & 12.54      & 46.08      & 37.05      \\
FGSM {\scriptsize $\epsilon\! =\! 0.031$}\hspace*{-4mm} & 8.94       & 30.84       & 36.31      & 52.80      \\
FGSM {\scriptsize $\epsilon\! =\! 0.062$}\hspace*{-4mm} & 10.62      & 30.28      & 36.78      & 51.60      \\
\bottomrule
\end{tabular}
\label{tab:adv_acc}
\end{table}

\subsection{Metric for statistical signatures}

Although we aim to identify inference time signatures of attacks, it is also interesting to explore the statistical signatures that our quantifications may exhibit over sets of clean and attacked samples. We thus employ the discrete version of the Bhattacharyya coefficient (BC) \cite{Bhattacharyya1946} for scalar-valued quantification, i.e., $\text{RF}$, $\text{PH}$, $\text{AD}$'s, and $S_\text{CKA}$. 

Let $h$ and $h'$ denote the empirical distributions (or histograms) of one of our quantifications with 100 bins computed for the sets of clean samples and attacked samples, respectively. Their $i$-th bin, $h_i$ and $h_i'$, give the numbers of samples that fall into that bin, normalized to sum up to one, where the bins are configured to cover the minimum and maximum values of the quantification. BC is defined as 
\begin{equation}
    BC(h, h') = \sum_i \sqrt{h_i h'_i}.
\end{equation}

The range of $\text{BC}$ is $[0, 1]$ where $\text{BC} = 1$ means the most similar. For better discrimination of clean and adversarial samples, we hope for a value as close as possible to 0, which gives a significant signature of adversarial attacks.
However, we realistically expect the value to be rather high for typical (and thus ``interesting'') configurations of methods for adversarial attacks, as only a small perturbation is introduced.

\section{Results}
\subsection{Frequency ratio}

\begin{figure}[t]
\begin{center}
\includegraphics[width=1.1\linewidth]{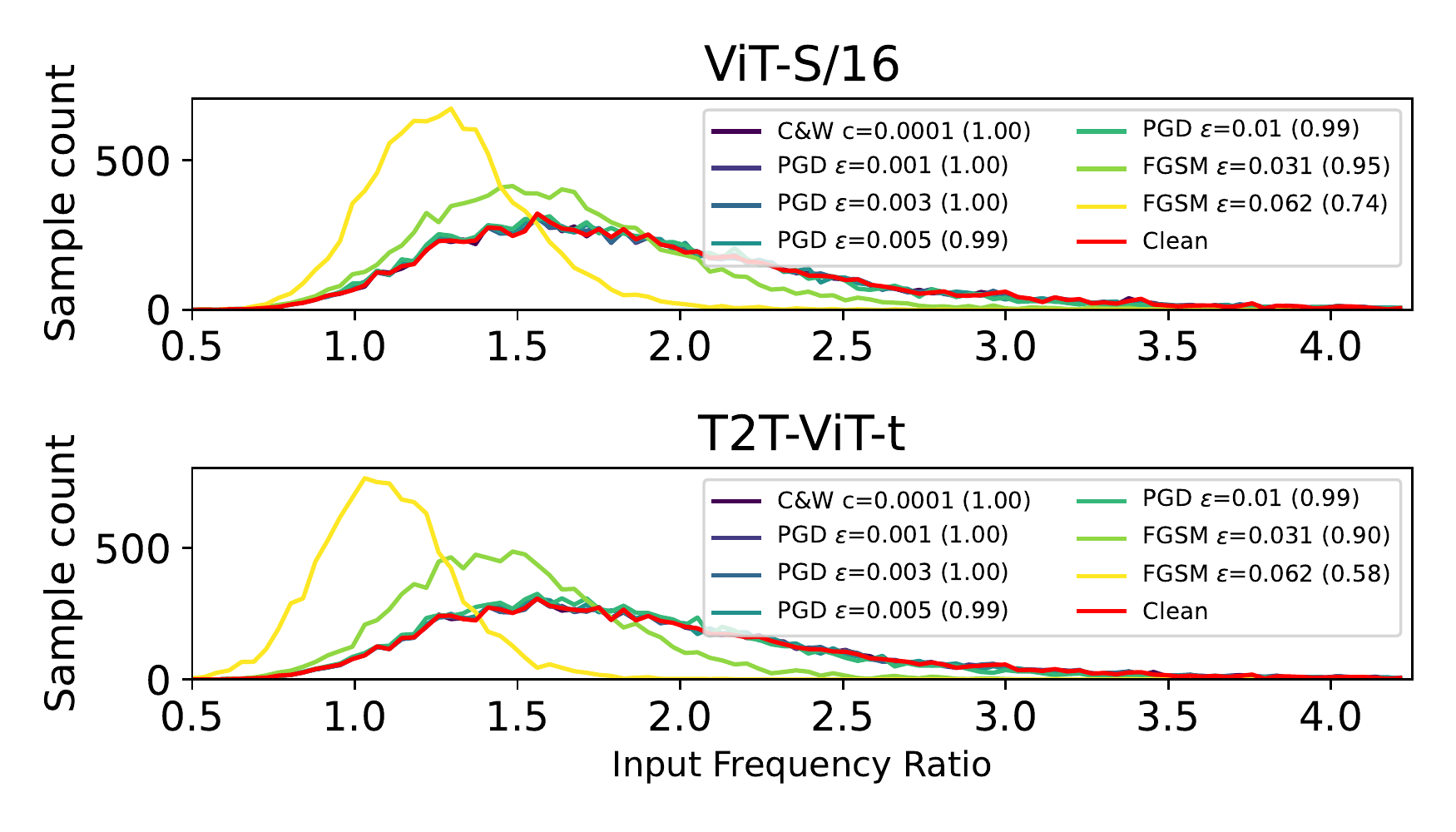}
\end{center}
\vspace{-0.5cm}
\caption{Distributions of $\text{FR}$ for representative two models. The numbers in Parentheses are BCs computed between the clean distribution and respective distributions with perturbations.}
\label{fig:Frequencies_valid}
\end{figure}

Figure \ref{fig:Frequencies_valid} shows the empirical distributions of $\text{FR}$ for clean samples and attacked samples by the configurations presented earlier. We can see the distributions are different only for the configurations with the largest perturbations  (namely FGSM with $\epsilon$ being 0.031 and 0.062), for which FR consistently drifts towards 1. This means that the energy of the perturbation $\eta$ is relatively concentrated in lower frequencies.
This confirms that ViTs are also vulnerable to low frequency when given enough perturbation budget \cite{Paul2021vision}.

The distributions for smaller perturbations are almost the same as the clean one, which is also shown in the high BCs computed for the attacked distributions; therefore, FR in itself cannot be reliable for attack detection.
We also note at this stage that the distribution shift is related to the perturbation strength and not to the success rate of the attack in Table \ref{tab:adv_acc}.
We will observe that same behavior on the other signatures and discuss this problem in the conclusion.
We know from \cite{spectralDefense2021} that it is possible to train detector models on Fourier Transform of an input to detect to presence of adversarial attack on CNNs, hence giving hint for similar results with ViTs if the same technique was to be used.
Analysis in the frequency domain can provide some signatures of attacks if the perturbation is large enough but is not useful otherwise.

\subsection{Posterior probability}

\begin{figure}[t]
\begin{center}
\includegraphics[width=1.1\linewidth]{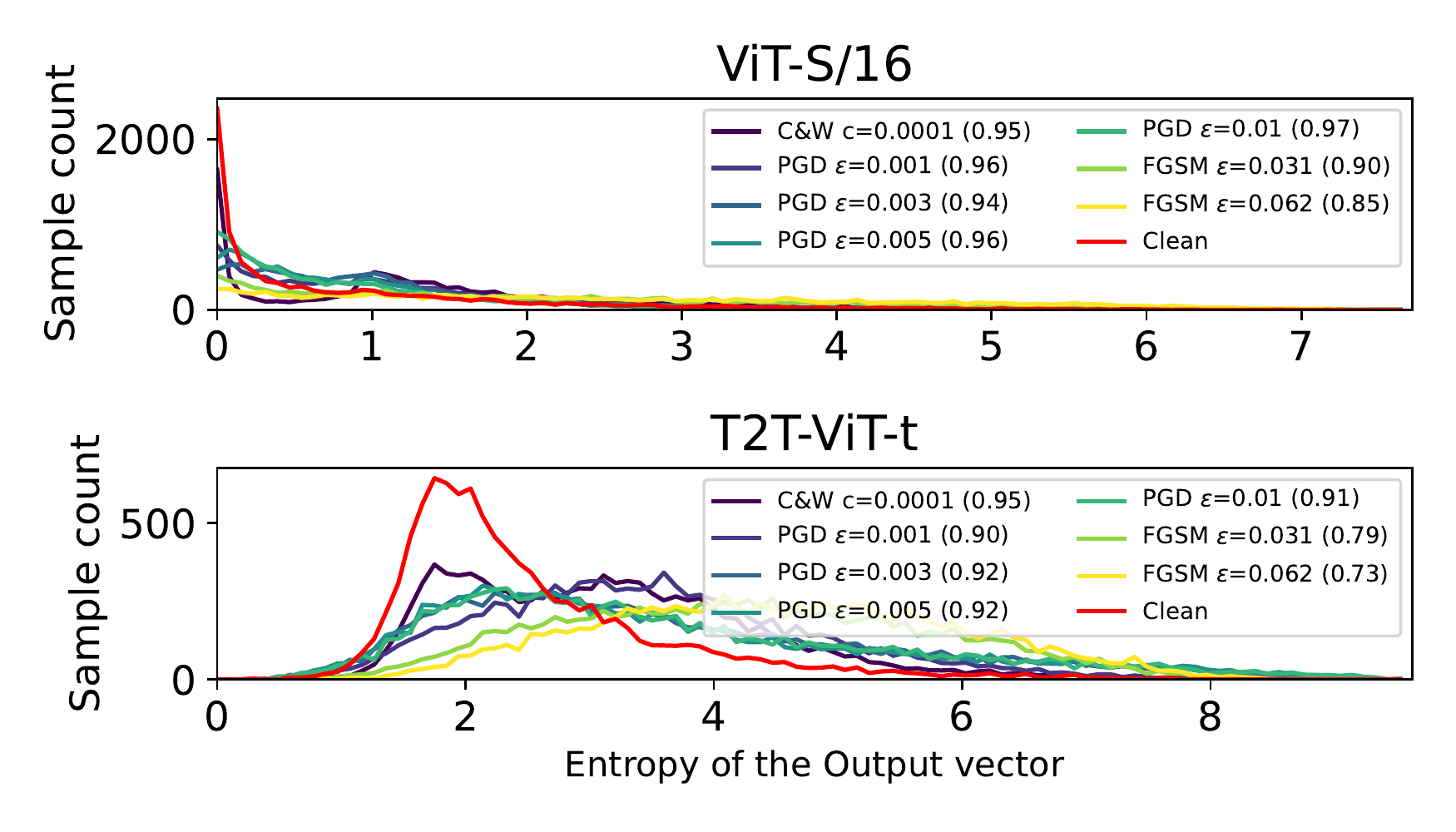}
\end{center}
\vspace{-0.5cm}
\caption{Distributions of $\text{PH}$ for the two models.}
\label{fig:Output_etpy_valid}
\end{figure}

The distributions of $\text{PH}$ are shown in Figure \ref{fig:Output_etpy_valid}.
We see very different behaviors from one model to another. The T2T-ViT model displays noticeably more varied distributions compared to the vanilla ViT model.
We cannot explain that difference considering that on average the T2T outperforms the ViT in the classification task.
For that matter the ViT even have better top-5 accuracy (96.13\%) on clean samples compared to the T2T (95.72\%).
This result demonstrates that $\text{PH}$ may be used in an attack detection pipeline for T2T-ViTs, though we would have to make heavy compromises in the accuracy or in the false positive rate if the quantification is used by itself.
It has to be noted that the significance of this quantification is also proportional to the intensity of the attack rather than the success rate of this one.

\subsection{Attention distances and profiles}
\label{sec:Att_dist}
In order to analyze the attention distance, we base our observations on the way the attention profile is computed in \cite{raghu2021vision}: the profile is computed for the entire test set, regardless of the accuracy of the detection.
This base gives us the reference of the nominal behavior of our model for in-distribution inputs.

Figure \ref{fig:Att_profile_ViT_T2T} shows the distributions of $S_\text{AP}(x)$. Note that we can still compute the distribution for the set of clean samples, which shows the difference between individual samples and their mean.

\begin{figure}[t]
\begin{center}
\includegraphics[width=1.1\linewidth]{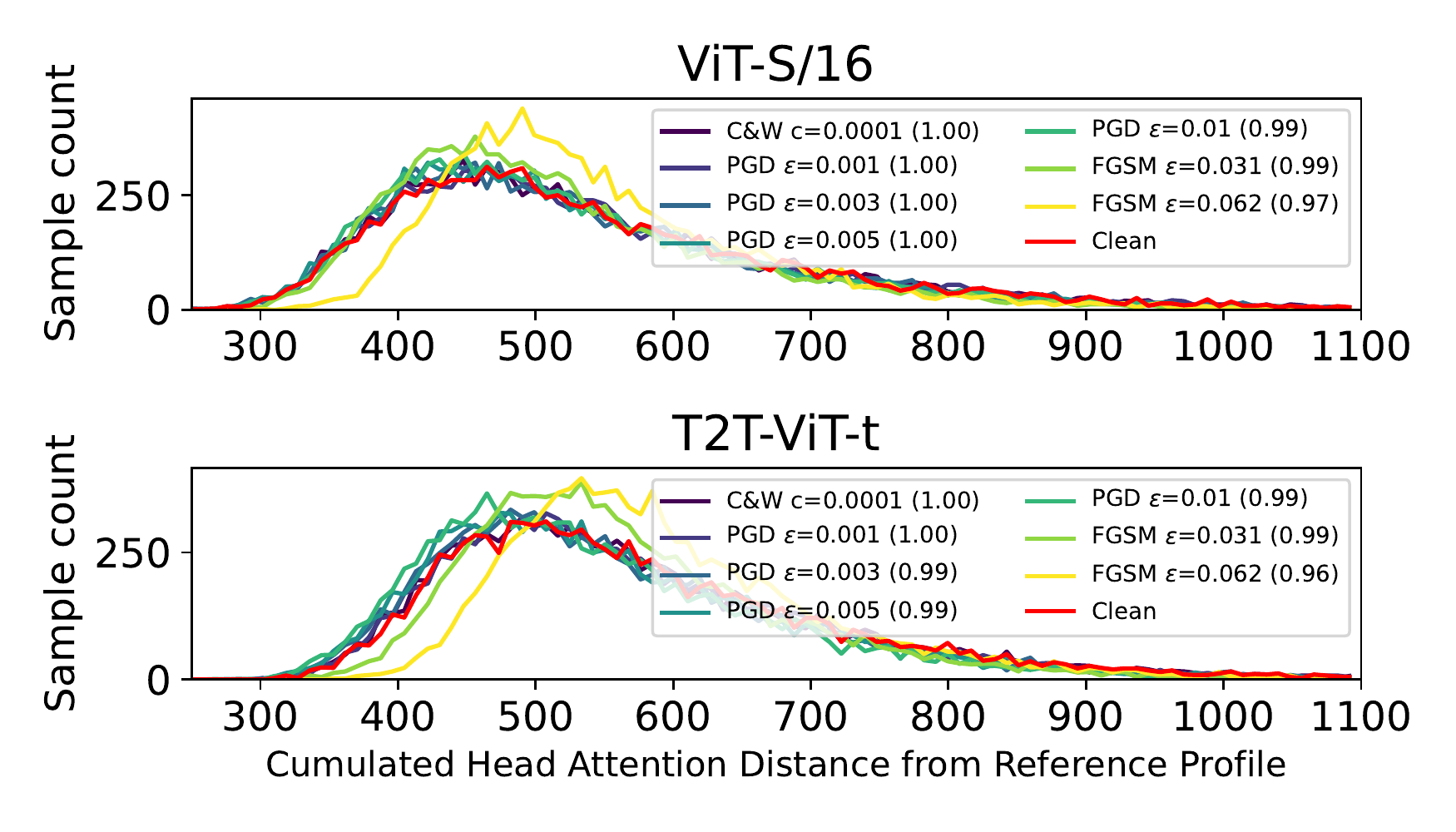}
\end{center}
\vspace{-0.5cm}
\caption{Distributions of $S_\text{AP}(x)$.}
\label{fig:Attention_valid}
\end{figure}

\begin{figure}[t]
\begin{center}
\includegraphics[width=\linewidth]{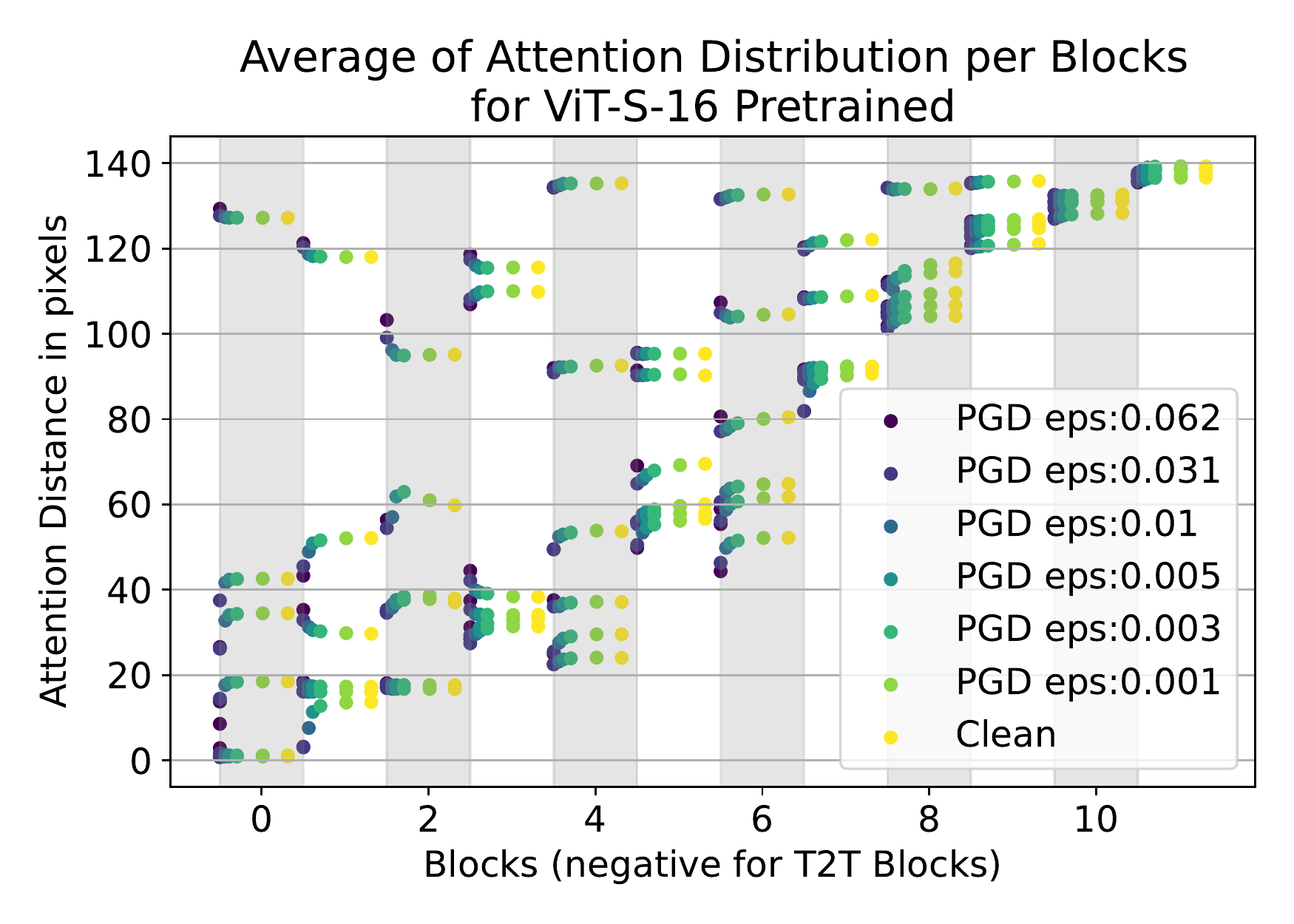}
\end{center}
\vspace{-0.5cm}
\caption{ADs per head observed on ViT-S/16 for PGD attacks. The different attacks are sorted so that the configuration that gives the lowest accuracy comes on the left within each head. Each point is the mean over the 10k samples.}
\label{fig:Attention_drift}
\end{figure}

We observe that $S_\text{AP}(x)$ behaves in a similar way as the previous metrics, showing a marginal separation ($\mathrm{BC} = 0.97$) for the largest perturbation on both models.
Since we measure the absolute difference between the reference and observed profiles, $S_\text{AP}(x)$ gives the gist but very little information on the actual direction of the change.
We can observe finer behavior when looking in detail at the level of attention heads.
In Figure \ref{fig:Attention_drift}, we show that perturbations either diversify or shrink ADs on several heads, mostly in the first blocks, and the trend of changes in ADs is constant for them, i.e., either diversified or shrunk, while all the others remain unaffected.
For instance, the head with the largest AD in block 1 is diversified along with the amount of perturbation, while the one with the second largest AD in the same block shrinks. 
We note that diversification mostly occurs in the lower half (i.e., smaller block indices) of the network while shrinkage happens throughout it.

One simple explanation in which the discrepancy in the attention results in a successful attack is that the perturbation distracts the attention of all patches to some regions that have no features leading to the ground-truth class. The attack may be successful if the features from irrelevant patches overwrite the features from the residual connection. From this argument, we hypothesized that heads with attention discrepancy are the sweet spot for adversarial attacks.
Verification of this hypothesis is up to further investigation.

\begin{table}[t]
\begin{center}
\caption{BCs computed for the most volatile heads in terms of attention distances.}
\label{tab:refined_att}
\begin{tabular}{lll}
\toprule
Attacks                                                        & ViT-S/16 & T2T-ViT-t \\
\midrule
C\&W $c=1\times 10^{-4}$                                                    & 1.00 (+0.00)     & 1.00 (+0.00)      \\
PGD $\epsilon = 1 \times 10^{-3}$                                            & 1.00 (+0.00)     & 1.00 (+0.00)      \\
PGD $\epsilon = 3 \times 10^{-3}$                                            & 1.00 (+0.00)     & 1.00 (+0.00)      \\
PGD $\epsilon = 5 \times 10^{-3}$                                            & 1.00 (+0.00)     & 1.00 (+0.00)      \\
PGD $\epsilon = 1 \times 10^{-2}$                                           & 0.99 (+0.00)     & 0.99 (+0.00)     \\
FGSM $\epsilon = 0.031$                                          & 0.97 (+0.02)     & 0.98 (+0.01)      \\
FGSM $\epsilon = 0.062$                                          & 0.85 (+0.12)     & 0.96 (+0.01)     \\
\bottomrule
\end{tabular}
\end{center}
\end{table}

We think those volatile heads are a good hook to build an attack detector in the future.
We designed $S_\text{AP}(x)$ to be a summary of all heads in all blocks, so the distributions in Figure ~\ref{fig:Attention_valid} may be too vague. 
Focusing on a few heads that exhibit large discrepancies can give a clearer distinction between clean and adversarial samples.
Table \ref{tab:refined_att} shows the BC values computed over $\text{AD}(x)$ for a cherry-picked head of each model and each configuration of attacks and the corresponding improvement of BC values.

\subsection{CKA Similarity}
\label{sec:cka}

\begin{figure}[t]
\begin{center}
\includegraphics[width=1.1\linewidth]{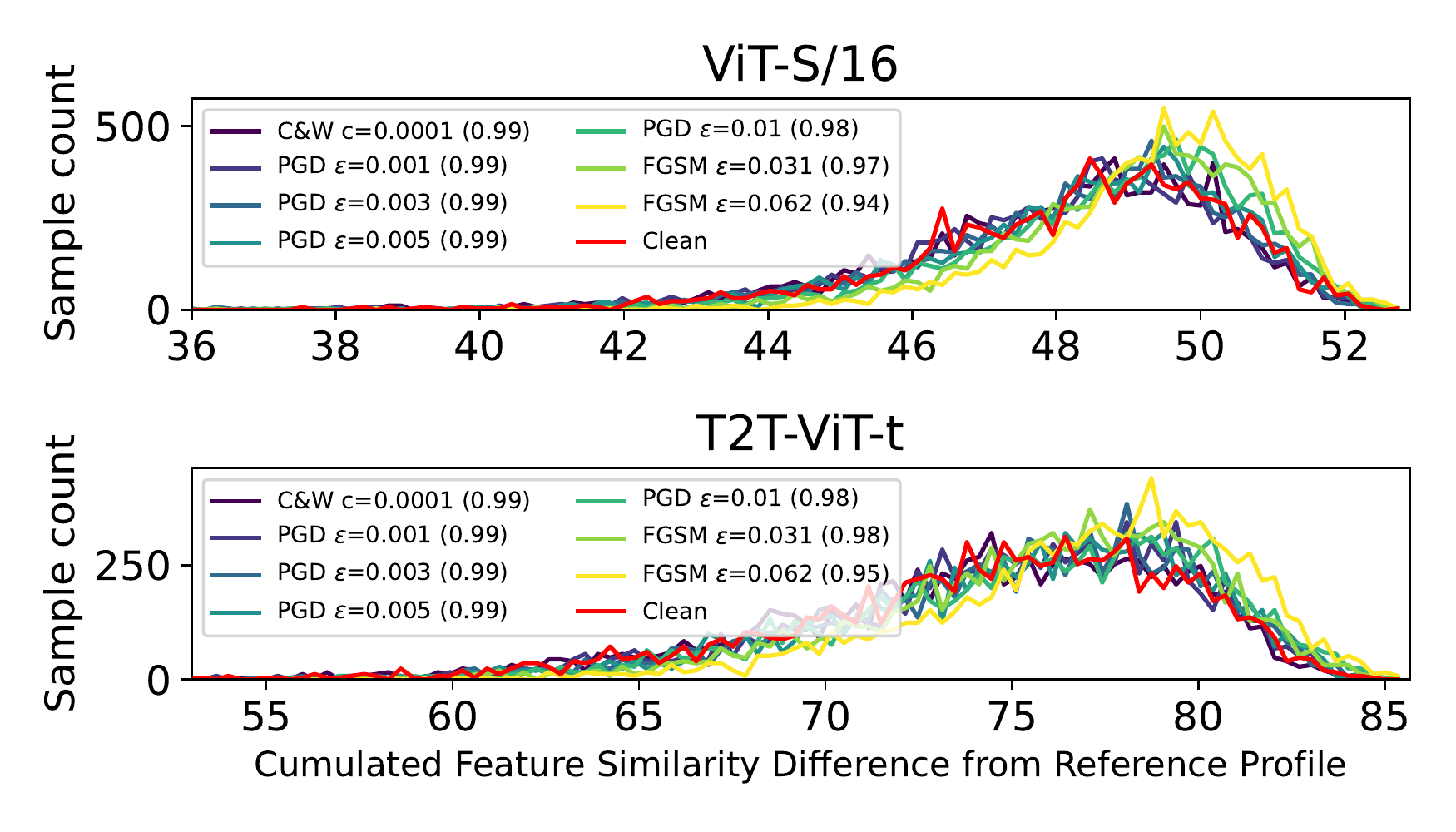}
\end{center}
\caption{Difference between the observed CKA similarity and the reference one.}
\label{fig:CKA_valid}
\end{figure}

The distributions of $S_\text{CKA}$ in \ref{fig:CKA_valid} show even worse separation capabilities compared to the other quantifications, where even the largest perturbations seem to have only slightly shifted the curves from the original clean data. 
In a similar fashion to AP, we take a step back from $S_\text{CKA}$ and look at the layer-wise difference $D_{ij}$ of the reference CKA similarity matrix and that of attacked samples, because the summary by $S_\text{CKA}$ does not provide sufficient information on the presence of any attacks.

\begin{figure}[t]
\begin{center}
\includegraphics[width=\linewidth]{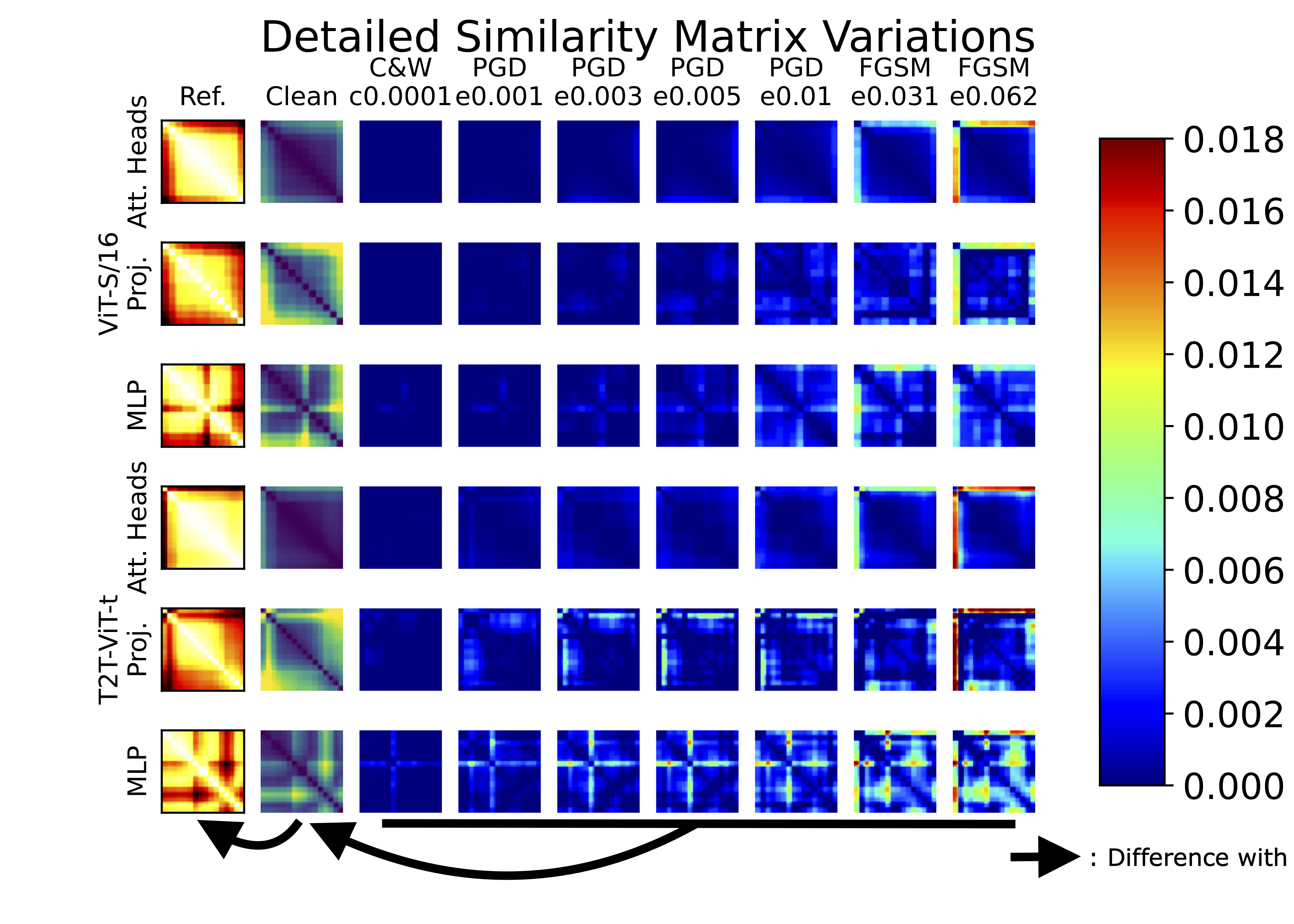}
\end{center}
\caption{Left-most column: Reference CKA similarity matrix $M^\text{ref}$. Second left column: $D$ computed for clean samples. Third left to right-most columns: $D$'s computed for samples under the attack of interest. From top to bottom: ViT-S/16's attention heads, projections, and MLPs; T2T-ViT-t's attention heads, projections, and MLPs.}
\label{fig:CKA_details}
\end{figure}

In Figure \ref{fig:CKA_details}, we observe clear patterns appearing with the increase of attack intensity.
The patterns are presumably related to the layers in Transformer blocks, which are comprised of layers after multi-head attention, after projection, and after multi-layer perceptron (MLP). 
One can point that the latent vectors after the attention heads are remarkably stable, and only the first layer (or two in the case of T2T-ViT) are affected when using the largest perturbations.
Ones after the projection layers are mildly affected, diffusing for all layers.  A remarkable impact is observed for our largest FGSM attack ($\epsilon=0.062$), where the first layer reacts strongly.
The last latent vectors are the one after MLPs of each attention block.
These vectors are affected throughout the whole spectrum of attacks, and the effect is noticeable around one specific block, that is, the 7-th one for ViT-S/16 and also the 7-th, which is the 5-th if we count in the ViT backbone only, for the T2T-ViT-t.
The perturbation seems to locally spread from this block with the intensity increased, as well as the perturbation appearing in the first blocks when using strong FGSM attacks.

\begin{table}[t]
\begin{center}
\caption{BCs computed for the most reactive latent vectors in terms of the CKA similarity.}
\label{tab:refined_CKA}
\begin{tabular}{lll}
\toprule
Attacks                                                        & ViT-S/16 & T2T-ViT-t \\
\midrule
C\&W $c = 1 \times 10^{-4}$                                                    & 0.99 (+0.00)     & 0.99 (+0.00)      \\
PGD $\epsilon = 1 \times 10^{-3}$                                            & 0.99 (+0.00)     & 0.99 (+0.00)      \\
PGD $\epsilon 3 \times 10^{-3}$                                            & 0.99 (+0.00)     & 0.98 (+0.01)      \\
PGD $\epsilon = 5 \times 10^{-3}$                                            & 0.99 (+0.00)     & 0.98 (+0.01)      \\
PGD $\epsilon = 1 \times 10^{-2}$                                            & 0.99 (-0.01)     & 0.98 (+0.00)     \\
FGSM $\epsilon = 0.031$                                          & 0.97 (+0.00)     & 0.96 (+0.02)      \\
FGSM $\epsilon = 0.062$                                          & 0.93 (+0.01)     & 0.91 (+0.04)     \\
\bottomrule
\end{tabular}
\end{center}
\end{table}

We show in Table \ref{tab:refined_CKA} the fine-grained BCs for cherry-picked layers (i.e. the most reactive layers) for each model.
The table shows that targeting specific layers can slightly help separate attacked samples from clean ones.
However, the separation is still marginal, and the majority of the samples are confused with the clean distribution.

\section{Conclusion}
We presented in this paper four different types of quantifications that potentially exhibit some signatures of adversarial attacks in the inference time.
The biggest challenge in adversarial attack detection comes from the great variability in models' reactions, and our experimental results generally show the difficulty in building a simple criterion to distinguish an attacked sample from clean ones.
We empirically demonstrated that the input frequency ratio and the posterior entropy are promising out-of-the-box.
We also showed that quantifications based on the latent vectors, such as attention profiles and CKA similarity allow separating only the largest attacks.
However, these quantifications render some patterns specific to each model, which may collectively serve as a stronger signature. 
The results also showed that these quantifications mostly react to the amplitude of perturbations but not to the effectiveness of the attack (Table \ref{tab:adv_acc}). This implies that fine-tuned adversarial attacks such as C\&W and PGD with a very small perturbation budget are both very efficient and undetectable. 
The current state of inference time signature is very rough, and the very little information does not seem worth the cost of the computation.
However we have great hope in the finer analysis of the latent vector to get better understanding in the future of where in the network will we have better chance of detecting an attack.
Future work includes exploring interactions between quantifications as stated above.

\section{Acknowledgements}
This work was partly supported by JST CREST Grant No. JPMJCR20D3.

\bibliography{aaai23}

\begin{thebibliography}{40}
\providecommand{\natexlab}[1]{#1}

\bibitem[{Albahar and Almalki(2019)}]{albahar2019deepfakes}
Albahar, M.; and Almalki, J. 2019.
\newblock Deepfakes: Threats and countermeasures systematic review.
\newblock \emph{Journal of Theoretical and Applied Information Technology},
  97(22): 3242--3250.

\bibitem[{Bai et~al.(2021)Bai, Mei, Yuille, and Xie}]{Bai2021Are}
Bai, Y.; Mei, J.; Yuille, A.~L.; and Xie, C. 2021.
\newblock Are Transformers more robust than {CNN}s?
\newblock \emph{Advances in Neural Information Processing Systems}, 34:
  26831--26843.

\bibitem[{Benz et~al.(2021)Benz, Zhang, Ham, Karjauv, and
  Kweon}]{benz2021robustness}
Benz, P.; Zhang, C.; Ham, S.; Karjauv, A.; and Kweon, I. 2021.
\newblock Robustness comparison of vision transformer and {MLP}-Mixer to
  {CNN}s.
\newblock In \emph{Proc. CVPR Workshop on Adversarial Machine Learning in
  Real-World Computer Vision Systems and Online Challenges (AML-CV)}, 21--24.

\bibitem[{Bhattacharyya(1946)}]{Bhattacharyya1946}
Bhattacharyya, A. 1946.
\newblock On a measure of divergence between two multinomial populations.
\newblock \emph{Sankhy{\=a}: the Indian Journal of Statistics}, 401--406.

\bibitem[{Biggio and Roli(2018)}]{Biggio_2018}
Biggio, B.; and Roli, F. 2018.
\newblock Wild patterns: Ten years after the rise of adversarial machine
  learning.
\newblock \emph{Pattern Recognition}, 84: 317--331.

\bibitem[{Carlini and Wagner(2017{\natexlab{a}})}]{carlini2017adversarial}
Carlini, N.; and Wagner, D. 2017{\natexlab{a}}.
\newblock Adversarial examples are not easily detected: Bypassing ten detection
  methods.
\newblock In \emph{Proc. 10th ACM workshop on artificial intelligence and
  security}, 3--14.

\bibitem[{Carlini and Wagner(2017{\natexlab{b}})}]{CWTowardsEval2016}
Carlini, N.; and Wagner, D. 2017{\natexlab{b}}.
\newblock Towards evaluating the robustness of neural networks.
\newblock In \emph{Proc. IEEE symposium on security and privacy}, 39--57. IEEE.

\bibitem[{Caron et~al.(2021)Caron, Touvron, Misra, J{\'e}gou, Mairal,
  Bojanowski, and Joulin}]{caron2021emerging}
Caron, M.; Touvron, H.; Misra, I.; J{\'e}gou, H.; Mairal, J.; Bojanowski, P.;
  and Joulin, A. 2021.
\newblock Emerging properties in self-supervised vision transformers.
\newblock In \emph{Proc. IEEE/CVF International Conference on Computer Vision},
  9650--9660.

\bibitem[{Chen et~al.(2022)Chen, Wang, Ming, and Feng}]{ChenViT-P2022}
Chen, B.; Wang, R.; Ming, D.; and Feng, X. 2022.
\newblock {ViT-P}: Rethinking Data-efficient Vision Transformers from Locality.
\newblock \emph{arXiv preprint arXiv:2203.02358}.

\bibitem[{Choromanski et~al.(2020)Choromanski, Likhosherstov, Dohan, Song,
  Gane, Sarlos, Hawkins, Davis, Mohiuddin, Kaiser
  et~al.}]{ChororethinhkPerf2020}
Choromanski, K.; Likhosherstov, V.; Dohan, D.; Song, X.; Gane, A.; Sarlos, T.;
  Hawkins, P.; Davis, J.; Mohiuddin, A.; Kaiser, L.; et~al. 2020.
\newblock Rethinking attention with performers.
\newblock \emph{arXiv preprint arXiv:2009.14794}.

\bibitem[{Cintas et~al.(2021)Cintas, Speakman, Akinwande, Ogallo, Weldemariam,
  Sridharan, and McFowland}]{CintasAttactDetect2020}
Cintas, C.; Speakman, S.; Akinwande, V.; Ogallo, W.; Weldemariam, K.;
  Sridharan, S.; and McFowland, E. 2021.
\newblock Detecting adversarial attacks via subset scanning of autoencoder
  activations and reconstruction error.
\newblock In \emph{Proc. International Conference on International Joint
  Conferences on Artificial Intelligence}, 876--882.

\bibitem[{Devlin et~al.(2018)Devlin, Chang, Lee, and
  Toutanova}]{devlin2019bert}
Devlin, J.; Chang, M.-W.; Lee, K.; and Toutanova, K. 2018.
\newblock Bert: Pre-training of deep bidirectional transformers for language
  understanding.
\newblock \emph{arXiv preprint arXiv:1810.04805}.

\bibitem[{Dosovitskiy et~al.(2010)Dosovitskiy, Beyer, Kolesnikov, Weissenborn,
  Zhai, Unterthiner, Dehghani, Minderer, Heigold, Gelly
  et~al.}]{dosovitskiy2021image}
Dosovitskiy, A.; Beyer, L.; Kolesnikov, A.; Weissenborn, D.; Zhai, X.;
  Unterthiner, T.; Dehghani, M.; Minderer, M.; Heigold, G.; Gelly, S.; et~al.
  2010.
\newblock An image is worth 16x16 words: Transformers for image recognition at
  scale.
\newblock \emph{arXiv preprint arXiv:2010.11929}.

\bibitem[{Fu et~al.(2022)Fu, Zhang, Wu, Wan, and Lin}]{fu2022patch}
Fu, Y.; Zhang, S.; Wu, S.; Wan, C.; and Lin, Y. 2022.
\newblock Patch-Fool: Are Vision Transformers Always Robust Against Adversarial
  Perturbations?
\newblock \emph{arXiv preprint arXiv:2203.08392}.

\bibitem[{Goodfellow, Shlens, and Szegedy(2014)}]{goodfellow2014explaining}
Goodfellow, I.~J.; Shlens, J.; and Szegedy, C. 2014.
\newblock Explaining and harnessing adversarial examples.
\newblock \emph{arXiv preprint arXiv:1412.6572}.

\bibitem[{Gupta, Dasgupta, and Akhtar(2020)}]{GuptaAdv2020}
Gupta, K.~D.; Dasgupta, D.; and Akhtar, Z. 2020.
\newblock Adversarial input detection using image processing techniques
  ({IPT}).
\newblock In \emph{2020 11th IEEE Annual Ubiquitous Computing, Electronics \&
  Mobile Communication Conference (UEMCON)}, 0309--0315. IEEE.

\bibitem[{Han et~al.(2022)Han, Wang, Chen, Chen, Guo, Liu, Tang, Xiao, Xu, Xu
  et~al.}]{han2022survey}
Han, K.; Wang, Y.; Chen, H.; Chen, X.; Guo, J.; Liu, Z.; Tang, Y.; Xiao, A.;
  Xu, C.; Xu, Y.; et~al. 2022.
\newblock A survey on vision transformer.
\newblock \emph{IEEE transactions on pattern analysis and machine
  intelligence}.

\bibitem[{Harder et~al.(2021)Harder, Pfreundt, Keuper, and
  Keuper}]{spectralDefense2021}
Harder, P.; Pfreundt, F.-J.; Keuper, M.; and Keuper, J. 2021.
\newblock Spectraldefense: Detecting adversarial attacks on {CNN}s in the
  fourier domain.
\newblock In \emph{2021 International Joint Conference on Neural Networks
  (IJCNN)}, 1--8. IEEE.

\bibitem[{Hendrycks and Gimpel(2016)}]{hendrycks2016early}
Hendrycks, D.; and Gimpel, K. 2016.
\newblock Early methods for detecting adversarial images.
\newblock \emph{arXiv preprint arXiv:1608.00530}.

\bibitem[{Kiefer and Wolfowitz(1952)}]{SGDKeifer1952}
Kiefer, J.; and Wolfowitz, J. 1952.
\newblock Stochastic estimation of the maximum of a regression function.
\newblock \emph{The Annals of Mathematical Statistics}, 462--466.

\bibitem[{Kornblith et~al.(2019)Kornblith, Norouzi, Lee, and
  Hinton}]{kornblith2019similarity}
Kornblith, S.; Norouzi, M.; Lee, H.; and Hinton, G. 2019.
\newblock Similarity of neural network representations revisited.
\newblock In \emph{International Conference on Machine Learning}, 3519--3529.
  PMLR.

\bibitem[{Li et~al.(2021)Li, Cao, Zhang, Chen, and Tan}]{li2021learning}
Li, J.; Cao, J.; Zhang, Y.; Chen, J.; and Tan, M. 2021.
\newblock Learning Defense Transformers for Counterattacking Adversarial
  Examples.
\newblock \emph{arXiv preprint arXiv:2103.07595}.

\bibitem[{Liu et~al.(2021)Liu, Lin, Cao, Hu, Wei, Zhang, Lin, and
  Guo}]{ze2021SWIN}
Liu, Z.; Lin, Y.; Cao, Y.; Hu, H.; Wei, Y.; Zhang, Z.; Lin, S.; and Guo, B.
  2021.
\newblock Swin transformer: Hierarchical vision transformer using shifted
  windows.
\newblock In \emph{Proc. IEEE/CVF International Conference on Computer Vision},
  10012--10022.

\bibitem[{Madry et~al.(2017)Madry, Makelov, Schmidt, Tsipras, and
  Vladu}]{TowardsMadry2017}
Madry, A.; Makelov, A.; Schmidt, L.; Tsipras, D.; and Vladu, A. 2017.
\newblock Towards deep learning models resistant to adversarial attacks.
\newblock \emph{arXiv preprint arXiv:1706.06083}.

\bibitem[{Mahmood, Mahmood, and Van~Dijk(2021)}]{MahmoodOnTheRobust2021}
Mahmood, K.; Mahmood, R.; and Van~Dijk, M. 2021.
\newblock On the robustness of vision transformers to adversarial examples.
\newblock In \emph{Proc. IEEE/CVF International Conference on Computer Vision},
  7838--7847.

\bibitem[{Mao et~al.(2022)Mao, Qi, Chen, Li, Duan, Ye, He, and
  Xue}]{towardsRobustMao2021}
Mao, X.; Qi, G.; Chen, Y.; Li, X.; Duan, R.; Ye, S.; He, Y.; and Xue, H. 2022.
\newblock Towards robust vision transformer.
\newblock In \emph{Proc. IEEE/CVF Conference on Computer Vision and Pattern
  Recognition}, 12042--12051.

\bibitem[{Naseer et~al.(2021)Naseer, Ranasinghe, Khan, Hayat, Shahbaz~Khan, and
  Yang}]{NaseerIntriguingViT2021}
Naseer, M.~M.; Ranasinghe, K.; Khan, S.~H.; Hayat, M.; Shahbaz~Khan, F.; and
  Yang, M.-H. 2021.
\newblock Intriguing properties of vision transformers.
\newblock \emph{Advances in Neural Information Processing Systems}, 34:
  23296--23308.

\bibitem[{Paul and Chen(2022)}]{Paul2021vision}
Paul, S.; and Chen, P.-Y. 2022.
\newblock Vision transformers are robust learners.
\newblock In \emph{Proc. AAAI Conference on Artificial Intelligence},
  volume~36, 2071--2081.

\bibitem[{Raghu et~al.(2021)Raghu, Unterthiner, Kornblith, Zhang, and
  Dosovitskiy}]{raghu2021vision}
Raghu, M.; Unterthiner, T.; Kornblith, S.; Zhang, C.; and Dosovitskiy, A. 2021.
\newblock Do Vision Transformers see like Convolutional Neural Networks?
\newblock \emph{Advances in Neural Information Processing Systems}, 34:
  12116--12128.

\bibitem[{Shao et~al.(2021)Shao, Shi, Yi, Chen, and
  Hsieh}]{shao2021adversarial}
Shao, R.; Shi, Z.; Yi, J.; Chen, P.-Y.; and Hsieh, C.-J. 2021.
\newblock On the adversarial robustness of vision transformers.
\newblock \emph{arXiv preprint arXiv:2103.15670}.

\bibitem[{Steiner et~al.(2021)Steiner, Kolesnikov, Zhai, Wightman, Uszkoreit,
  and Beyer}]{SteinerHowToTrain2021}
Steiner, A.; Kolesnikov, A.; Zhai, X.; Wightman, R.; Uszkoreit, J.; and Beyer,
  L. 2021.
\newblock How to train your {ViT}? data, augmentation, and regularization in
  vision transformers.
\newblock \emph{arXiv preprint arXiv:2106.10270}.

\bibitem[{Sun et~al.(2017)Sun, Shrivastava, Singh, and
  Gupta}]{sun2017revisiting}
Sun, C.; Shrivastava, A.; Singh, S.; and Gupta, A. 2017.
\newblock Revisiting unreasonable effectiveness of data in deep learning era.
\newblock In \emph{Proc. IEEE international conference on computer vision},
  843--852.

\bibitem[{Suwajanakorn, Seitz, and
  Kemelmacher-Shlizerman(2017)}]{suwajanakorn2017synthesizing}
Suwajanakorn, S.; Seitz, S.~M.; and Kemelmacher-Shlizerman, I. 2017.
\newblock Synthesizing obama: learning lip sync from audio.
\newblock \emph{ACM Transactions on Graphics (ToG)}, 36(4): 1--13.

\bibitem[{Szegedy et~al.(2013)Szegedy, Zaremba, Sutskever, Bruna, Erhan,
  Goodfellow, and Fergus}]{goodfellow2013intriguing}
Szegedy, C.; Zaremba, W.; Sutskever, I.; Bruna, J.; Erhan, D.; Goodfellow, I.;
  and Fergus, R. 2013.
\newblock Intriguing properties of neural networks.
\newblock \emph{arXiv preprint arXiv:1312.6199}.

\bibitem[{Touvron et~al.(2021)Touvron, Cord, Douze, Massa, Sablayrolles, and
  J{\'e}gou}]{Touvron2020DEIT}
Touvron, H.; Cord, M.; Douze, M.; Massa, F.; Sablayrolles, A.; and J{\'e}gou,
  H. 2021.
\newblock Training data-efficient image transformers \& distillation through
  attention.
\newblock In \emph{International Conference on Machine Learning}, 10347--10357.
  PMLR.

\bibitem[{Uelwer, Michels, and Candido(2021)}]{UelwerLearning2021}
Uelwer, T.; Michels, F.; and Candido, O.~D. 2021.
\newblock Learning to Detect Adversarial Examples Based on Class Scores.
\newblock In \emph{German Conference on Artificial Intelligence (K{\"u}nstliche
  Intelligenz)}, 233--240. Springer.

\bibitem[{Vaswani et~al.(2017)Vaswani, Shazeer, Parmar, Uszkoreit, Jones,
  Gomez, Kaiser, and Polosukhin}]{vaswani2017attention}
Vaswani, A.; Shazeer, N.; Parmar, N.; Uszkoreit, J.; Jones, L.; Gomez, A.~N.;
  Kaiser, {\L}.; and Polosukhin, I. 2017.
\newblock Attention is all you need.
\newblock \emph{Advances in neural information processing systems}, 30.

\bibitem[{Wang et~al.(2022)Wang, Bai, Zhou, and Xie}]{wang2022can}
Wang, Z.; Bai, Y.; Zhou, Y.; and Xie, C. 2022.
\newblock Can {CNN}s Be More Robust Than Transformers?
\newblock \emph{arXiv preprint arXiv:2206.03452}.

\bibitem[{Yuan et~al.(2021)Yuan, Chen, Wang, Yu, Shi, Jiang, Tay, Feng, and
  Yan}]{yuan2021tokenstotoken}
Yuan, L.; Chen, Y.; Wang, T.; Yu, W.; Shi, Y.; Jiang, Z.-H.; Tay, F.~E.; Feng,
  J.; and Yan, S. 2021.
\newblock Tokens-to-token {ViT}: Training vision transformers from scratch on
  imagenet.
\newblock In \emph{Proc. IEEE/CVF International Conference on Computer Vision},
  558--567.

\bibitem[{Zantedeschi, Nicolae, and Rawat(2017)}]{zantedeschi2017efficient}
Zantedeschi, V.; Nicolae, M.-I.; and Rawat, A. 2017.
\newblock Efficient defenses against adversarial attacks.
\newblock In \emph{Proc. 10th ACM Workshop on Artificial Intelligence and
  Security}, 39--49.

\end{thebibliography}
\newpage
\phantom{Necessary to make the full page jump.}
\newpage
\section{Supplementary Material}

\subsection{Frequency ratio}

\begin{figure}[h]
\begin{center}
\includegraphics[width=\linewidth]{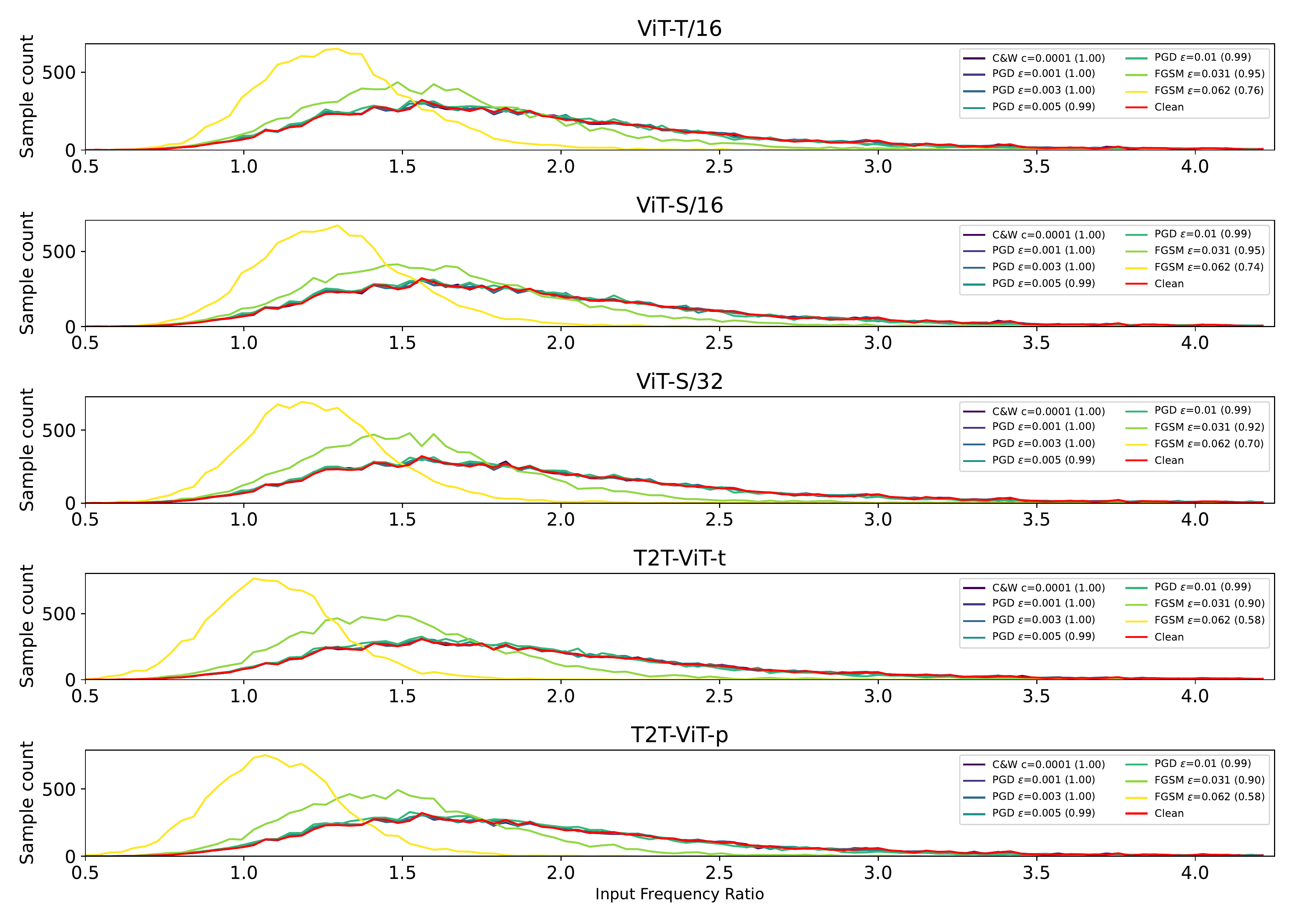}
\end{center}
\caption{Distributions of $\text{FR}$ for representative two models. The numbers in Parentheses are BCs computed between the clean distribution and respective distributions with perturbations.}
\label{fig:FR_extended}
\end{figure}

We present in Figure \ref{fig:FR_extended} the distribution over 3 additional network variations as presented in Table \ref{tab:model-tab}.
We clearly see a direct extension of the result observed on ViT-S/16 reproduced in both the smaller ViT-T/16 and the ViT-S/32 with larger patch size.
We note a small variation of the BC on the ViTs for the largest FGSM attack, but we think it is within the margin of error.
As for the T2T family the results here are exactly similar, that is in complete agreement with the performer being faster equivalent operation of the transformer.

\begin{table}[b]
\begin{center}
\caption{BCs computed for the most volatile heads in terms of attention distances.}
\begin{tabular}{llllll}
\toprule
Attacks                                                        & ViT-T/16 & ViT-S/16 & ViT-S/32 & T2T-ViT-t & T2T-ViT-p \\
\midrule
C\&W $c=1\times 10^{-4}$                                        & 1.00 (+0.00) & 1.00 (+0.00) & 1.00 (+0.00)    & 1.00 (+0.00) & 1.00 (+0.00) \\
PGD $\epsilon = 1 \times 10^{-3}$                               & 1.00 (+0.00) & 1.00 (+0.00) & 1.00 (+0.00)    & 1.00 (+0.00) & 1.00 (+0.00) \\
PGD $\epsilon = 3 \times 10^{-3}$                               & 0.99 (+0.01) & 1.00 (+0.00) & 1.00 (-0.01)    & 1.00 (+0.00) & 1.00 (-0.01) \\
PGD $\epsilon = 5 \times 10^{-3}$                               & 0.99 (+0.00) & 1.00 (+0.00) & 0.99 (+0.00)    & 1.00 (+0.00) & 1.00 (-0.01) \\
PGD $\epsilon = 1 \times 10^{-2}$                               & 0.99 (+0.00) & 0.99 (+0.00) & 0.99 (+0.00)    & 0.99 (+0.00) & 1.00 (-0.01) \\
FGSM $\epsilon = 0.031$                                         & 0.99 (+0.00) & 0.97 (+0.02) & 0.98 (+0.01)    & 0.98 (+0.01) & 0.99 (+0.00) \\
FGSM $\epsilon = 0.062$                                         & 0.98 (+0.00) & 0.85 (+0.12) & 0.99 (-0.05)    & 0.96 (+0.01) & 0.97 (+0.03) \\
\bottomrule
\end{tabular}
\end{center}
\label{tab:ext_refined_att}
\end{table}

\subsection{Posterior probability}

\begin{figure}[h]
\begin{center}
\includegraphics[width=\linewidth]{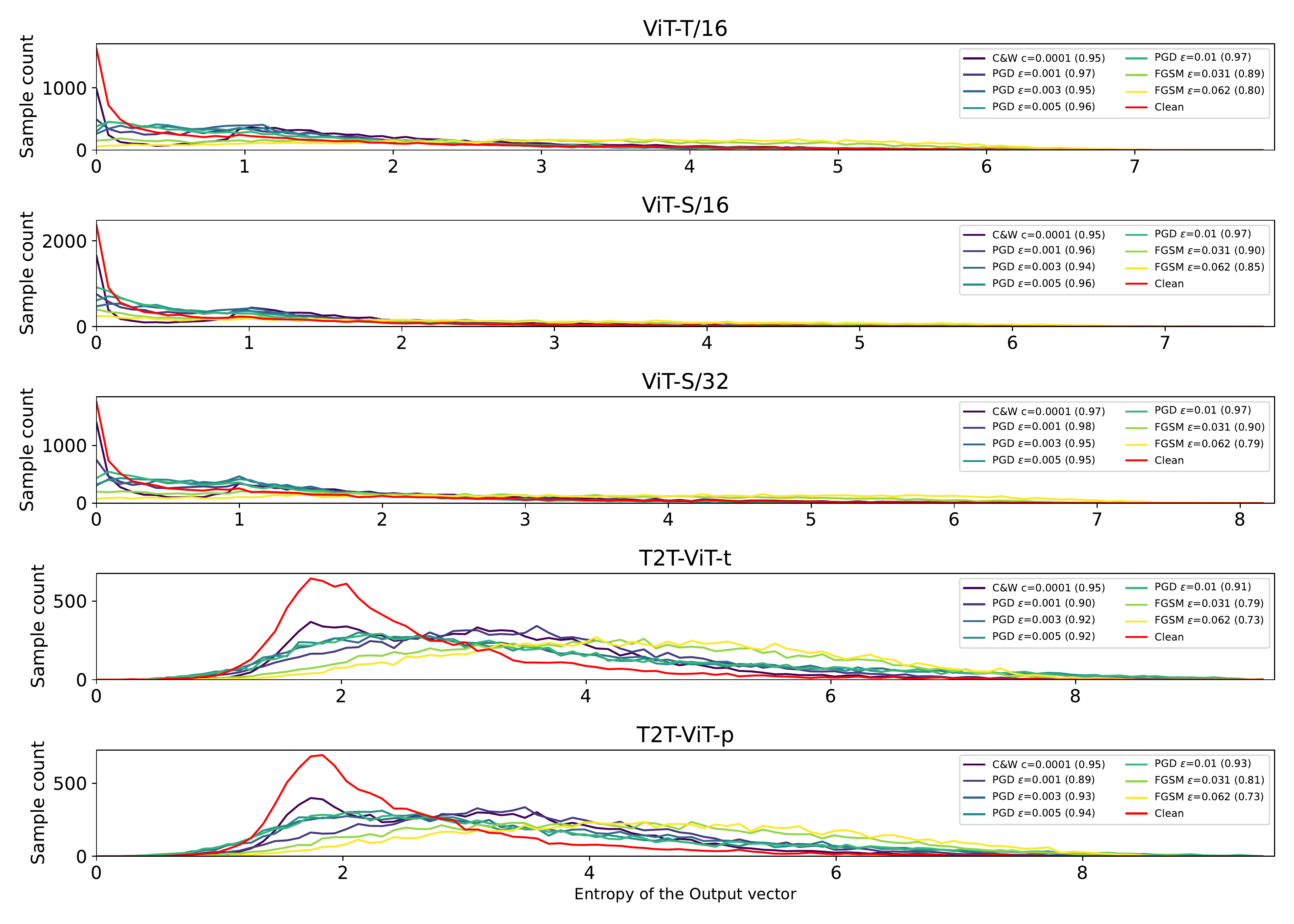}
\end{center}
\caption{Distributions of $\text{PH}$ for the two models.}
\label{fig:PH_extended}
\end{figure}

In Figure \ref{fig:PH_extended} we extend the evaluation to the 3 other models.
We show here that the behavior display by the ViT-S/16 and the T2T-ViT-t are completely representative of their family.
We find interesting here that the T2T-ViTs in general have entropy of approx. 2 while the ViTs are stuck at ~0.
The fact that the ViTs learned to output exactly one very high value as prediction is maintained in the adversarial setups, while the T2Ts who usually are more varied have even less decisive answers when attacked.
However we note that this behavior from the T2T family does not impact the accuracy and is even outperforming the ViTs. (Table \ref{tab:adv_acc}).

\newpage
\subsection{Attention distances and profiles}
\label{sec:Att_dist_ext}
\begin{figure}[h]
\begin{center}
\includegraphics[width=\textwidth]{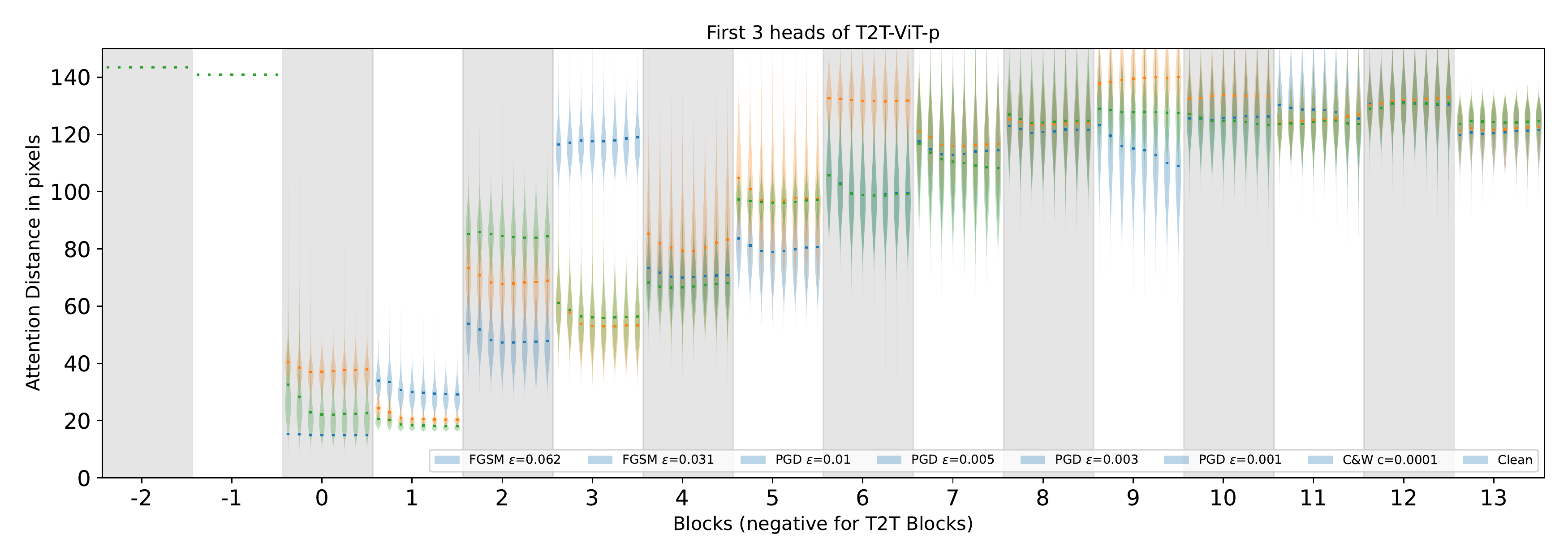}
\end{center}
\caption{Distribution of the ADs over the 10k samples for the first 3 heads observed on T2T-ViT-p for our set of attacks. The different attacks are sorted for each block in the same order as the legend and in Figure \ref{fig:Attention_drift}.}
\label{fig:violin_t2t-p}
\end{figure}

We also expend here the experiments of Section \ref{sec:Att_dist} and show here varied behaviors.
At first the BC computed for the ViTs increase in the order presented here with ViT-T having the lowest separability on the attention distance due to the low number of heads to compare.
The other ViT-T/32 shows better BC for the largest attacks. We attribute this to the fact that this network has by design a larger attention scope, and that making it diverge, only from one patch, is much more noticeable.
The T2Ts also present interesting changes, with the T2T-ViT-p showing very poor BC.
When we compare Figure \ref{fig:violin_t2t-t} and Figure \ref{fig:violin_t2t-p} we see a very clear difference in the attention of the T2T block where most of the drift under attacks were measured.

We also try to refine those scores with the same technique and observe very similar to the ones in their respective family again.

\begin{figure}[h]
\begin{center}
\includegraphics[width=\linewidth]{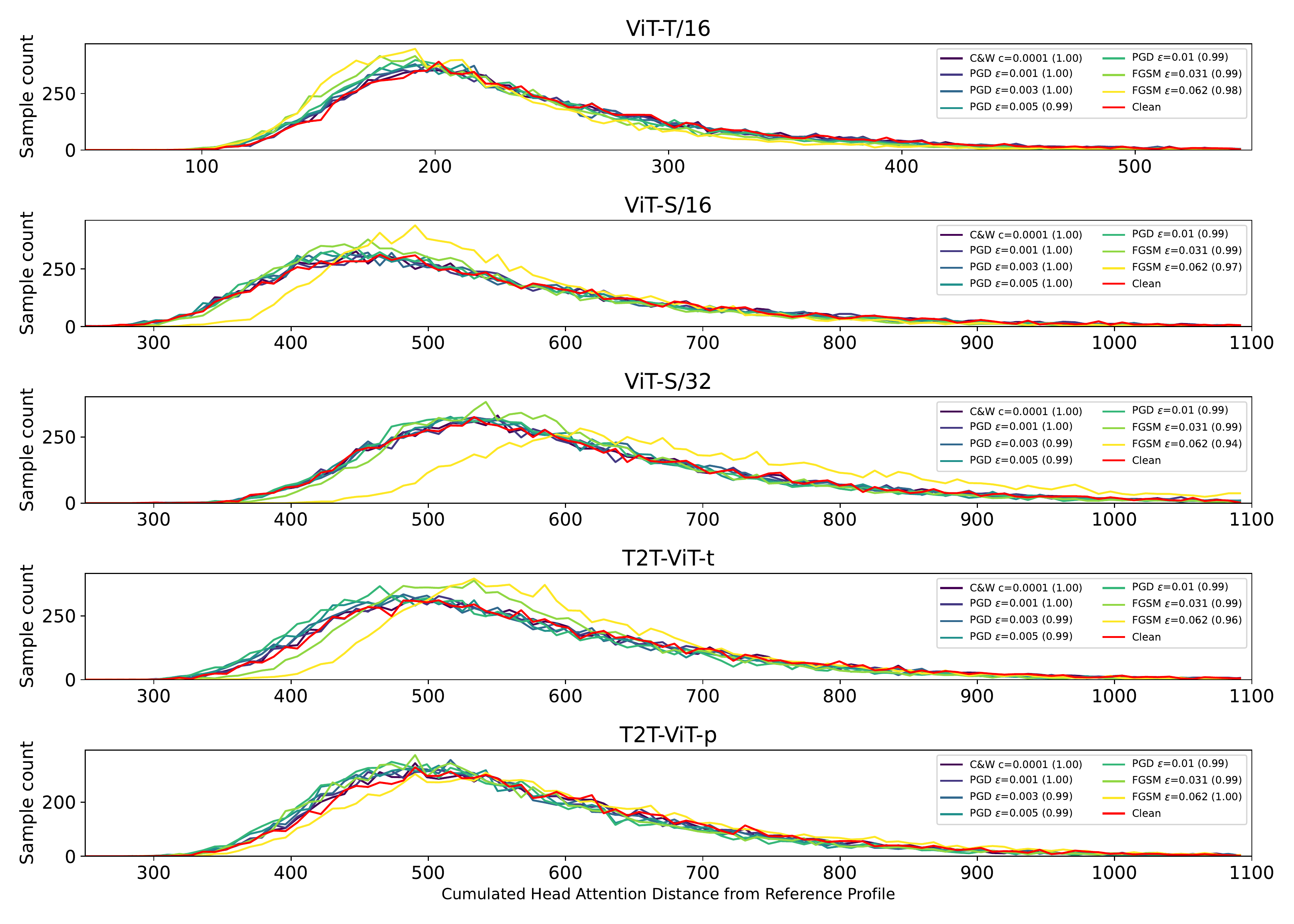}
\end{center}
\caption{ADs per head observed on ViT-S/16 for PGD attacks. The different attacks are sorted so that the configuration that gives the lowest accuracy comes on the left within each head. Each point is the mean over the 10k samples.}
\end{figure}

\subsection{Stability of the Attention distances and profiles}

\begin{figure}[h]
\begin{center}
\includegraphics[width=\textwidth]{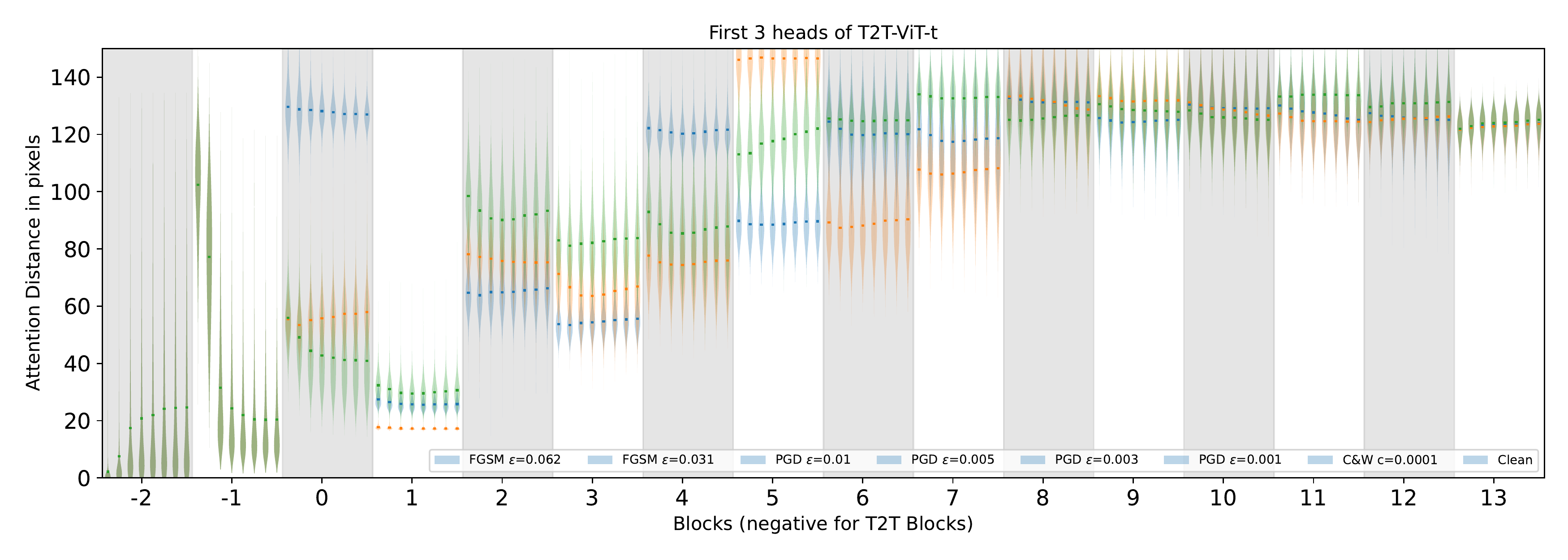}
\end{center}
\caption{Distribution of the ADs over the 10k samples for the first 3 heads observed on T2T-ViT-t for our set of attacks. The different attacks are sorted for each block in the same order as the legend and in Figure \ref{fig:Attention_drift}.}
\label{fig:violin_t2t-t}
\end{figure}

We show in this section some testimonies proving that the attention drifts reported in Section \ref{sec:Att_dist} are consistent.
We mainly show here that when the distributions are drifting, they do so in a very consistent way with no huge distribution spread or grouping (with the exception of the attention head -1 on Figure \ref{fig:violin_t2t-t}), meaning that making observations on the average is indeed representative of the drift measured.

\newpage
\phantom{Necessary to make the full page jump.}
\newpage
\newpage
\phantom{Necessary to make the full page jump.}

\begin{figure}[h]
\begin{center}
\includegraphics[width=\textwidth]{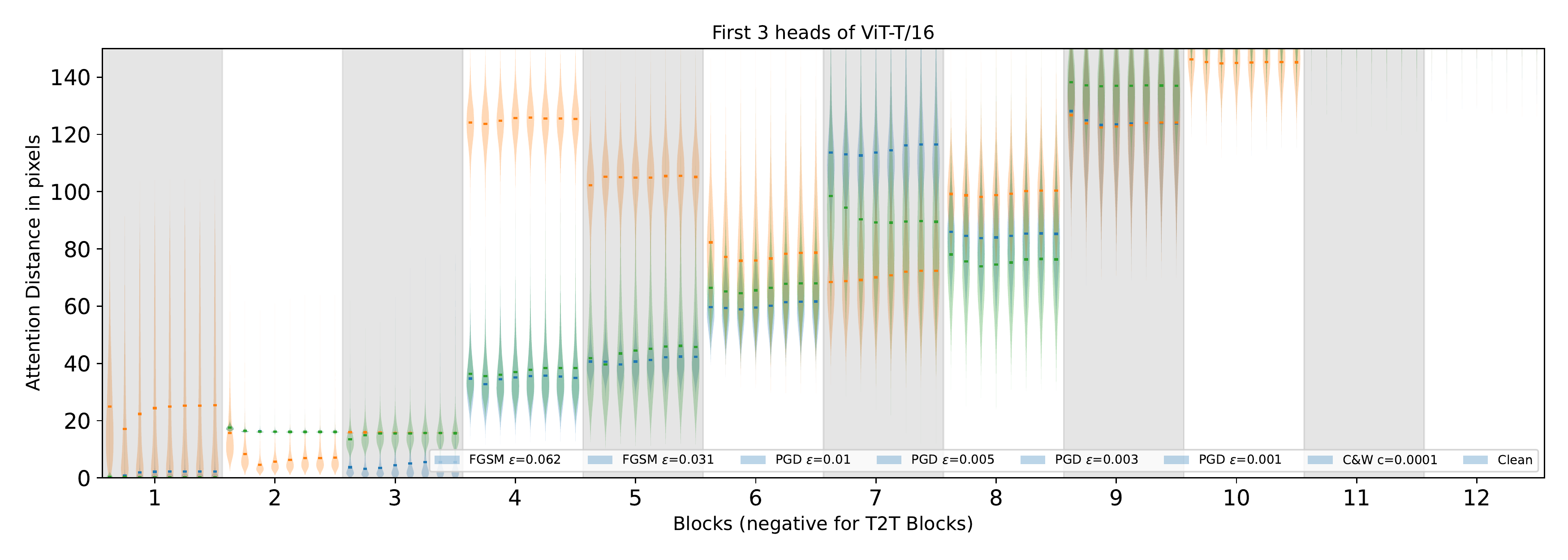}
\end{center}
\caption{Distribution of the ADs over the 10k samples for the first 3 heads observed on ViT-T/16 for our set of attacks. The different attacks are sorted for each block in the same order as the legend and in Figure \ref{fig:Attention_drift}.}
\end{figure}

\begin{figure}[h]
\begin{center}
\includegraphics[width=\textwidth]{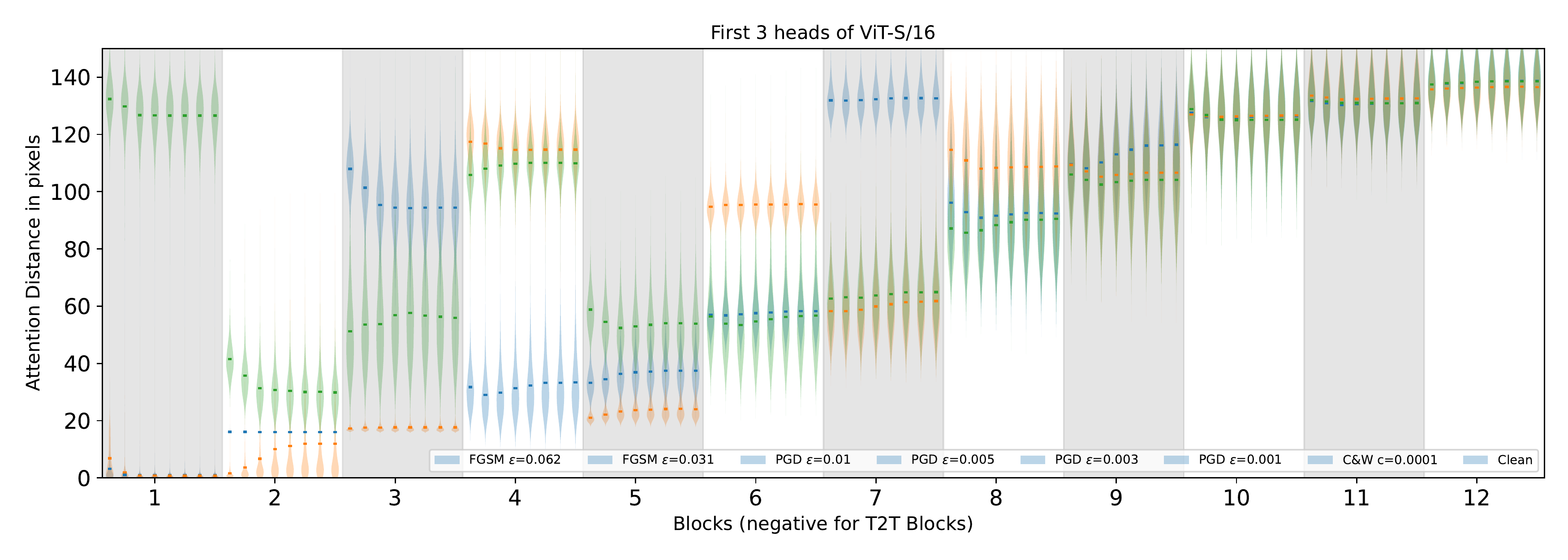}
\end{center}
\caption{Distribution of the ADs over the 10k samples for the first 3 heads observed on ViT-S/16 for our set of attacks. The different attacks are sorted for each block in the same order as the legend and in Figure \ref{fig:Attention_drift}.}
\end{figure}

\newpage
\phantom{Necessary to make the full page jump.}
\newpage
\newpage
\phantom{Necessary to make the full page jump.}

\begin{figure}[h]
\begin{center}
\includegraphics[width=\textwidth]{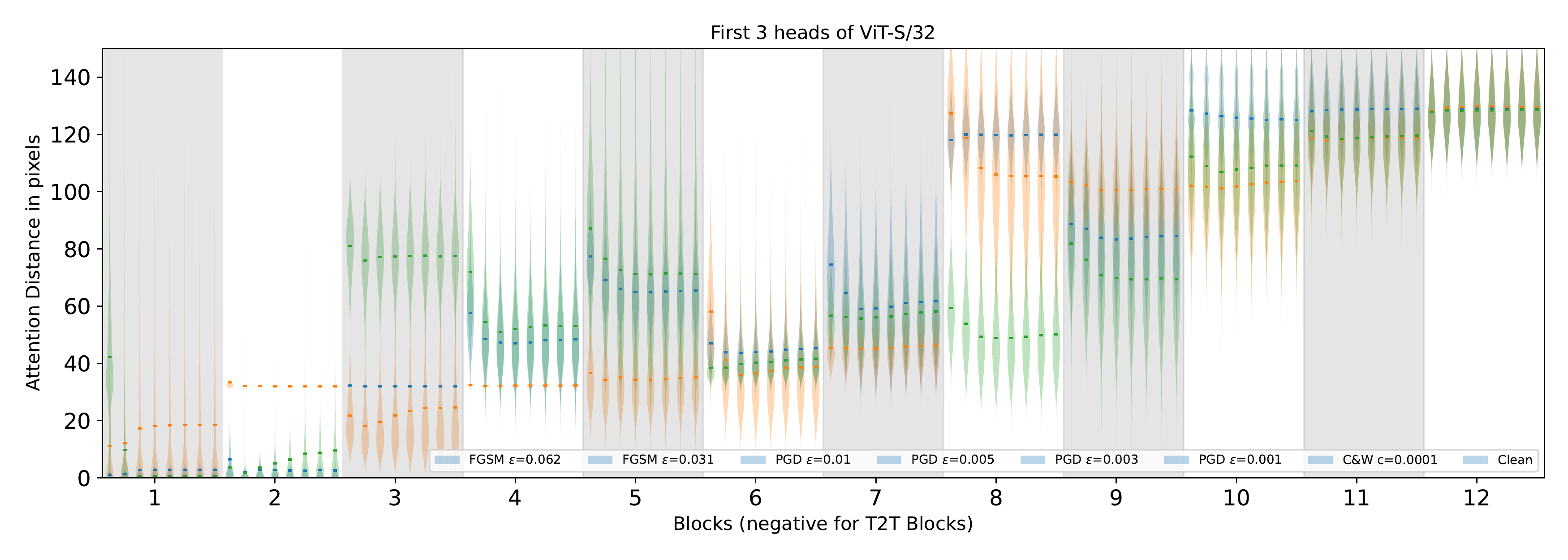}
\end{center}
\caption{Distribution of the ADs over the 10k samples for the first 3 heads observed on ViT-S/32 for our set of attacks. The different attacks are sorted for each block in the same order as the legend and in Figure \ref{fig:Attention_drift}.}
\end{figure}

\newpage
\phantom{Necessary to make the full page jump.}
\newpage
\newpage
\phantom{Necessary to make the full page jump.}

\end{document}